\documentclass[10pt,twocolumn,letterpaper]{article}

\usepackage[pagenumbers]{iccv} 
\usepackage{multirow}
\definecolor{iccvblue}{rgb}{0.21,0.49,0.74}
\definecolor{steelblue}{HTML}{4682B4}
\definecolor{lightcoral}{HTML}{F08080}
\usepackage[pagebackref,breaklinks,colorlinks,allcolors=iccvblue]{hyperref}
\newcommand{\myparagraph}[1]{{\vspace{.2em} \noindent \bf #1}}

\newcommand{\ieno}{\textit{i.e.}}

\def\paperID{*****} 
\def\confName{ICCV}
\def\confYear{2025}

\title{Why Compress What You Can Generate? \\When GPT-4o Generation Ushers in Image Compression Fields}

\author{
Yixin Gao\thanks{Equal contribution.} \quad
Xiaohan Pan\footnotemark[1] \quad
Xin Li\thanks{Xin Li and Zhibo Chen are corresponding authors.} \quad
Zhibo Chen\footnotemark[2] \\
\textit{University of Science and Technology of China} \\
{\tt\small \{gaoyixin, pxh123\}@mail.ustc.edu.cn, \{xin.li, chenzhibo\}@ustc.edu.cn}
}

\begin{document}
\maketitle
\begin{abstract}
The rapid development of AIGC foundation models has revolutionized the paradigm of image compression, which paves the way for the abandonment of most pixel-level transform and coding, compelling us to ask: \textbf{why compress what you can generate} if the AIGC foundation model is powerful enough to faithfully generate intricate structure and fine-grained details from nothing more than some compact descriptors, \ieno, texts, or cues. Fortunately, recent GPT-4o image generation of OpenAI has achieved impressive cross-modality generation, editing, and design capabilities, which motivates us to answer the above question by exploring its potential in image compression fields. In this work, we investigate two typical compression paradigms: textual coding and multimodal coding (\ieno, text + extremely low-resolution image), where all/most pixel-level information is generated instead of compressing via the advanced GPT-4o image generation function. The essential challenge lies in how to maintain semantic and structure consistency during the decoding process. To overcome this, we propose a structure raster-scan prompt engineering mechanism to transform the image into textual space, which is compressed as the condition of GPT-4o image generation. Extensive experiments have shown that the combination of our designed structural raster-scan prompts and GPT-4o's image generation function achieved the impressive performance compared with recent multimodal/generative image compression at ultra-low bitrate, further indicating the potential of AIGC generation in image compression fields.
\end{abstract}
    
\section{Introduction}
\label{sec:intro}

Lossy image compression has been an irreplaceable technology in visual signal communication, significantly  reducing bandwidth and storage costs. Currently, lossy image compression methods can be categorized into two types: i) Traditional image coding, where each module is manually designed with multiple modes, and rate-distortion optimization is performed to select the optimal mode~\cite{vvc,bellard_bpg,li2021task}. ii) Learned image coding, which uses deep neural network modules to implement nonlinear transform coding~\cite{balle2020nonlinear}, achieving end-to-end rate-distortion optimization~\cite{balle2018variational,cheng2020learned,guo2021causal,he2022elic,liu2023learned,minnen2018joint,minnen2020channel,feng2023nvtc,wu2021learned}. Both of them rely on pixel-wise transform coding to reduce the spatial redundancies. However, pixel-wise coding inevitably suffers from the inherent challenges: the semantic integrity and perceptual quality will be catastrophically compromised when the coding bitrates are constrained, especially for ultra-low bitrates.


In recent years, Artificial Intelligence-Generated Content (AIGC) has achieved remarkable progress, particularly in cross-modality image generation, demonstrating remarkable fidelity and alignment with conditional inputs (\eg, text, sketch maps, reference images). This breakthrough has paved the way for abandoning all/partial pixel-wise compression by raising the question: if the original pixels can be generated perfectly after coding, why compress what we can generate? Some previous works~\cite{lei2023text,perco,xue2025dlf,jia2024generative,diffeic,mao2024extreme,xue2024unifying,li2024misc} have offered valuable insights into this question. These studies explored leveraging the priors from pretrained generative models, \eg, Stable Diffusion~\cite{latentdiffusion} and VQ-GAN~\cite{vqgan}, to achieve high-fidelity image compression by transmitting only a small amount of highly compact representations, such as text, sketches, and compact latents. While promising, these approaches often require additional fine-tuning, which compromises the generality and flexibility of the generative model.
The recent breakthroughs in OpenAI’s GPT-4o model~\cite{openai2025gpt4o}, showcasing impressive multimodal image generation and editing capabilities, have generated significant excitement in the community. Motivated by this progress, we aim to explore whether GPT-4o can more effectively address the challenges outlined above without any additional training.

In this paper, we explore two compression paradigms: textual coding and multimodal coding (text combined with an extremely low-resolution image), where all/most pixel-level information is generated rather than compressed, leveraging the advanced image generation capabilities of GPT-4o. 
A major challenge in this generative compression paradigm is how to simultaneously preserve both perceptual quality and consistency with the original image during image generation. To overcome this, we propose a structural raster-scan prompt engineering method. Rather than producing a general caption of the original image, we design a prompt that explicitly guides the multimodal large language model (MLLM) to preserve structural layout and multi-level visual consistency with the original image. 
Specifically, the prompt instructs the MLLM to describe visual elements in a top-to-bottom, left-to-right raster-scan order, preserving the spatial arrangement and relative positioning of objects, which supports geometric consistency and structural integrity. 
In addition, it encourages descriptions that capture distinctive features, object identities, semantic roles, and style properties. By aligning with these multiple levels of visual information, the structured prompt enables more faithful image reconstructions under the generative compression paradigm. We find that this approach effectively maintains the relative positioning between objects, significantly improving the consistency of reconstruction image with the original image.

Building on this strategy, we achieve image compression at bitrates as low as 0.001 bpp while maintaining good perceptual quality and consistency. Experiments demonstrate that our method achieves competitive performance compared to recent generative compression approaches at ultra-low bitrates, even
without any additional training. These results underscore the promising potential of leveraging powerful AIGC foundation models in the field of image compression.

\section{Related Works}
\label{sec:related_works}

\subsection{Image Compression at Ultra-low Bitrates}
Compressing images at ultra-low bitrates (e.g., below 0.01 bpp) poses significant challenges due to the severely limited amount of information that can be retained. Most image codecs, which are typically optimized for distortion metrics such as PSNR, tend to produce blurry and unrealistic reconstructions when the bitrate is drastically reduced. To overcome this, recent approaches have turned to generative approaches that leverage learned visual priors to synthesize plausible image content, thereby enhancing perceptual quality even under extreme compression scenarios.

Current solutions for ultra-low bitrate generative compression can be broadly categorized into two representative paradigms: I) tokenizer-based frameworks that exploit compact latent representations; and ii) diffusion-based frameworks, which models fine-grained spatial details using iterative refinement. Tokenizer-based frameworks typically rely on image tokenizers with discrete codebooks, such as VQGAN~\cite{vqgan}, which learn semantically meaningful tokens via latent clustering. Several studies have employed such tokenizers to implement image codecs. Mao~\cite{mao2024extreme} modified the codebook size using K-means clustering in a pretrained VQGAN. Moreover, further bitrate reduction is achieved by discarding less informative tokens and regenerating them using transformers~\cite{xue2024unifying}. Jia~\cite{jia2024generative} enhanced performance through transform coding in the latent space of VQGAN. DLF~\cite{xue2025dlf} introduces a dual-branch latent fusion approach with cross-stream interactions, aligning semantic and detail components to achieve better adaptability and generation fidelity.

Diffusion-based frameworks, on the other hand, leverage the strong priors of diffusion models to enhance reconstruction fidelity. For example, Text+Sketch~\cite{lei2023text} utilizes Stable Diffusion (SD) to build a compression pipeline conditioned on both text prompts and sketch maps. PerCo~\cite{perco} integrates a fine-tuned diffusion model into its codec, combining vector-quantized spatial features and global text descriptions, and achieves impressive reconstruction quality at extremely low bitrates (below 0.01 bpp). MISC~\cite{li2024misc} introduces an ultra-low bitrate image compression framework that leverages large multimodal models to extract semantic information and spatial mappings to guide the diffusion process, enabling high-quality reconstruction of both natural and AI-generated images while achieving significant bitrate reduction. 
Recently, methods like DiffEIC~\cite{diffeic} and RDEIC~\cite{rdeic} use VAEs to compress image latents that guide diffusion models, further boosting performance.


\subsection{Conditional Image Generation}
Conditional image generation has emerged as a highly active research area in recent years, driven by significant advancements in generative modeling. The development of foundation models, including generative adversarial networks~\cite{goodfellow2014generative} and diffusion models~\cite{ho2020denoising,song2020score,li2023diffusion}, has significantly improved the fidelity of generated images. Among them, Stable Diffusion~\cite{latentdiffusion} stands out as a widely used latent diffusion model, featuring a latent U-Net architecture with cross-attention conditioning to generate high-resolution images from compact latent representations.

Conditional image generation typically involves various input modalities, \eg, text, image, sketch, audio, with text and images being the two most representative and extensively studied. 
Text-to-image generation, where a generative model synthesizes an image based on natural language description, has recently shown remarkable photorealism and strong semantic alignment with textual prompt~\cite{betker2023improving,esser2024scaling,flux2024,xie2025sana}. Similarly, image-to-image generation synthesizes a target image conditioned on an input image, enabling applications such as controllable image generation~\cite{controlnet,mou2024t2i,epstein2023diffusion}, style transfer~\cite{zhang2023inversion,wang2023stylediffusion,zhang2024artbank}, and image editing~\cite{kawar2023imagic,Mokady2023NullText,shi2024dragdiffusion,feng2025dit4edit}.

In particular, the recently introduced GPT-4o image generation~\cite{openai2025gpt4o} capability has demonstrated impressive performance in high-fidelity multimodal generation. In this paper, we explore the potential of GPT-4o's powerful cross-modal generation capabilities for image coding. We find that even without any additional training, our method achieves impressive performance compared with recent multimodal/generative image compression at ultra-low bitrates.
\section{Method}
\begin{figure*}[tb]
\centerline{\includegraphics[width=0.8\linewidth]{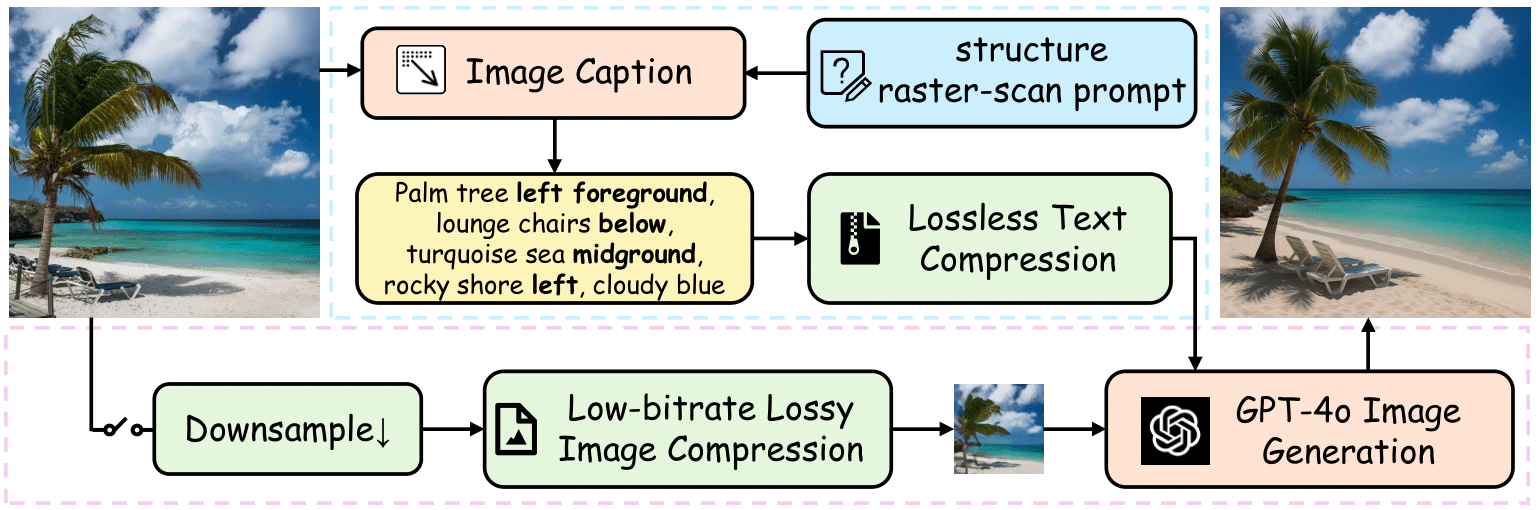}}
    \caption{Overall pipeline of our multimodal image compression framework based on GPT-4o image generation.}
    \vspace{-5mm}
\label{fig:pipeline}    
\end{figure*}

\subsection{Framework}
We propose a multimodal image compression framework that leverages the powerful image generation capabilities of GPT-4o. Unlike most existing generative codecs that require fine-tuning AIGC foundation models, our approach effectively harnesses GPT-4o’s multimodal generation ability in a plug-and-play manner, without the need for specialized model training. 
As illustrated in Fig.~\ref{fig:pipeline}, our framework uses two types of conditioning inputs: text and image, where the image branch can be optionally transmitted depending on bitrate requirements. 

Specifically, the original image is first captioned by the multimodal large language model using our proposed structurald raster-scan prompt, extracting text that contains both the semantic content of the main items and their spatial positions. This text is then losslessly compressed and transmitted to provide semantic guidance for the following image generation process.
The second branch, visual condition compression, transmits an extremely compressed image to offer base structural and color information for the generation process. Finally, GPT-4o generates the reconstructed image guided by the transmitted conditioning inputs. Next, we will introduce the two branch in detail.

\subsection{Structural Raster-scan Prompt}
With the rapid advancement of AIGC foundation models, text-to-image (T2I) generation has seen remarkable improvements. Models such as Stable Diffusion 3~\cite{esser2024scaling}, Flux~\cite{flux2024}, and DALL·E 3~\cite{dalle3} have demonstrated the ability to produce high-resolution, semantically rich, and stylistically diverse images from natural language descriptions. 
In particular, GPT-4o~\cite{openai2025gpt4o} showcases unprecedented multimodal capabilities, including accurate text rendering, detailed scene synthesis, and image editing guided by both textual and visual inputs. 
Therefore, when the prompt length is fixed (\ie, the compression cost remains constant), the specific content of the prompt plays a crucial role in determining the quality of the generated image. This raises a crucial question: how can we obtain text prompts that not only produce high perceptual quality but also preserve consistency with the original image? 

\begin{figure}[t]
    \centering
    \includegraphics[width=\linewidth]{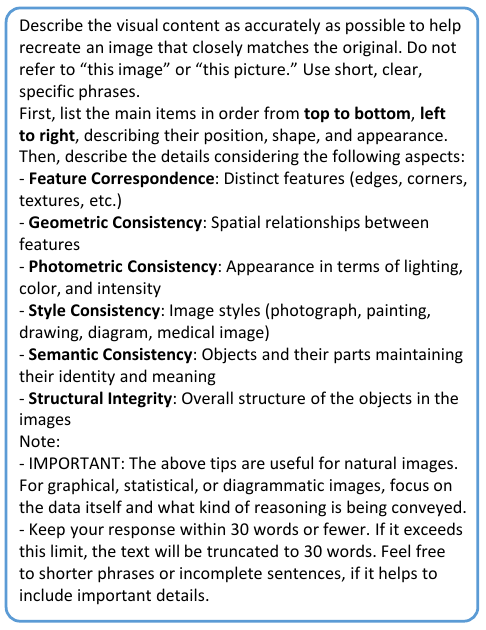}
    \caption{Our proposed structural raster-scan prompt. 
    }
    \label{fig:prompt1}
\end{figure}

To achieve this, as illustrated in Fig.~\ref{fig:prompt1}, we introduce a structural raster-scan prompt engineering approach. This prompt is carefully designed to guide multimodal large language models (MLLMs) in preserving the structural layout and maintaining multi-level visual consistency with the original image. Specifically, the prompt instructs the model to describe visual elements in a top-to-bottom, left-to-right sequence, emphasizing the position, shape, and appearance of objects. To further enhance consistency, we consider six dimensions that guide the model to elaborate on image content more precisely: feature correspondence captures local visual details such as textures and edges; geometric consistency describes spatial layout and object alignment; photometric properties refer to lighting, shadows, and color tones; stylistic alignment specifies the overall visual style (e.g., photo, painting, anime); semantic coherence ensures object completeness and plausibility; and structural integrity maintains the global composition. 
This comprehensive approach ensures that the spatial arrangement and stylistic attributes of visual elements are faithfully captured. 

Additionally, the prompt provides tailored instructions for different image types. By enforcing a concise format-limiting responses to a fixed number of words (\eg, 30 words) or fewer, it encourages clear, content-rich descriptions. This structurald methodology effectively supports high-fidelity image reconstruction within the generative compression framework.

\begin{figure*}[t]
    \centering
    \includegraphics[width=\linewidth]{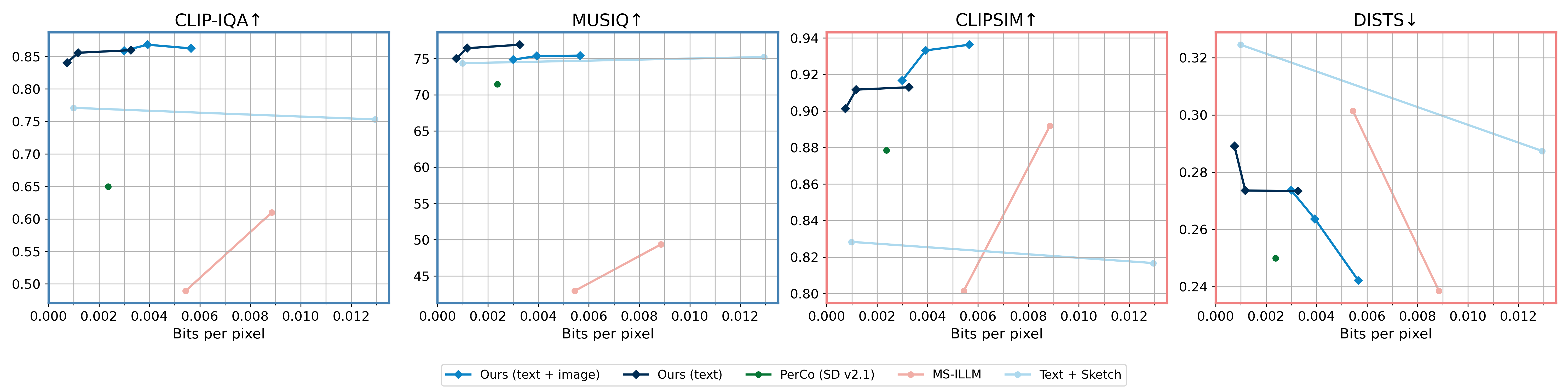}
    \caption{Tradeoffs between bitrate and different metrics on DIV2K. The quality is evaluated by both perceptual (\textcolor{steelblue}{blue frames}) and consistency metrics (\textcolor{lightcoral}{red frames}).}
    \label{fig:main_results}
\end{figure*}

\begin{figure*}[t]
    \centering
    \includegraphics[width=\linewidth]{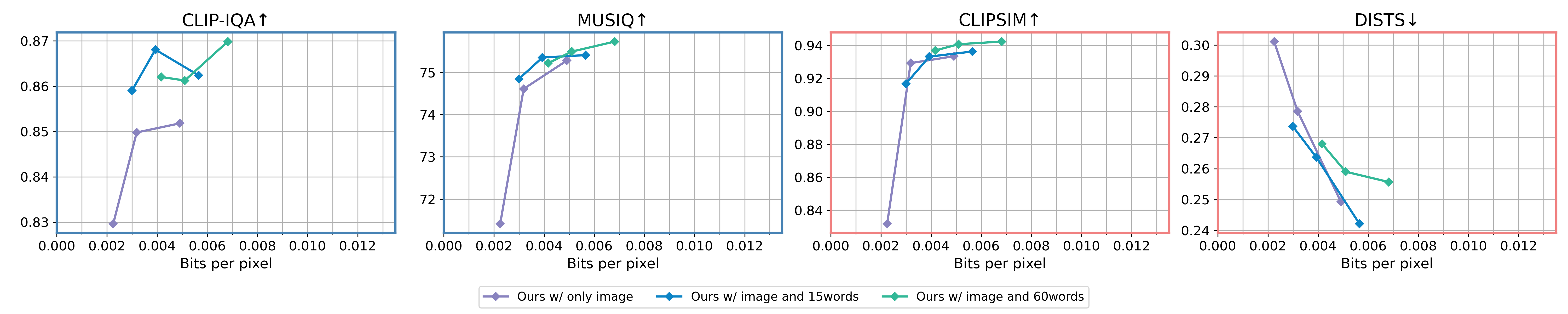}
    \caption{Ablation on prompt length.}
    \label{fig:Ablation_prompt_length}
\end{figure*}

\subsection{Visual Condition Compression}
The goal of the visual condition is to provide GPT-4o image generation with strong visual priors, ensuring that the reconstructed output remains faithful to the original image. 
A key challenge lies in identifying appropriate forms of visual priors. While visual cues such as sketches can assist in guiding image generation and offer good perceptual quality, they often suffer from severe consistency issues~\cite{lei2023text}. In contrast, compressed images retain much of the original structural and color information, which helps improve consistency scores. 
Following prior works~\cite{hoogeboom2023high,li2024misc,gao2024unimic,ghouse2023residual}, we adopt the strategy of using an extremely compressed image as the visual prior. Specifically, we downsample the input image by a factor of 8 to reduce redundancy, then apply low-bitrate compression using an image codec. 
To provide GPT-4o with a high-quality visual starting point, we employ the state-of-the-art perceptual codec MS-ILLM~\cite{msillm}, which effectively preserves key visual features while mitigating distortions such as blurriness and color shifts often introduced by MSE-optimized codecs. Finally, by feeding the compressed image along with textual prompt into GPT-4o, we enable a reconstruction process that achieves both high perceptual quality and high consistency.



\section{Experimental Results}
\subsection{Experiment Setup}
\myparagraph{Implementation details}\quad
We use Lempel-Ziv coding as implemented in the zlib library~\cite{zlib}, similar to Lei \etal\cite{lei2023text} and PerCo~\cite{perco}, to losslessly compress the text prompt and image size information. 
For image compression, we first downsample the original images by a factor of 8 and then compress them using the state-of-the-art GAN-based perceptual image codec, MS-ILLM~\cite{msillm}.

We test GPT-4o using custom automation scripts developed based on~\cite{gptimgeval}, which directly interact with the GPT-4o web interface. 
Considering the randomness of GPT-4o's image generation, all experiments are repeated three times, and the average is reported. 

\myparagraph{Datasets and Evaluation}\quad
For evaluation, we use the first 10 images from the DIV2K~\cite{div2k} validation set. Each image is resized so that its shorter side equals 1024px, and then center-cropped into a square shape of $1024\times1024$ pixels, following the approach of Yang \etal~\cite{yang2024lossy}.

To assess image quality, we consider both consistency and perceptual metrics. For consistency, we employ CLIPSIM~\cite{clip} as used in Text + Sketch~\cite{lei2023text}, which calculates the cosine similarity between CLIP embeddings to measure semantic similarity. Additionally, we use DISTS~\cite{dists}, a full-reference image quality metric that combines structural and texture similarity by comparing deep feature representations of the reference and distorted images.
For perception, the widely used metrics CLIP-IQA~\cite{clipiqa} and MUSIQ~\cite{ke2021musiq} are adopted. CLIP-IQA evaluates human aesthetic satisfaction with the image, while MUSIQ assesses both aesthetic and technical quality using a multi-scale transformer architecture.

\begin{figure*}[htbp]
    \centering

    \begin{minipage}{0.198\textwidth} 
        \centering
        \textbf{Origin}
        \includegraphics[width=\textwidth]{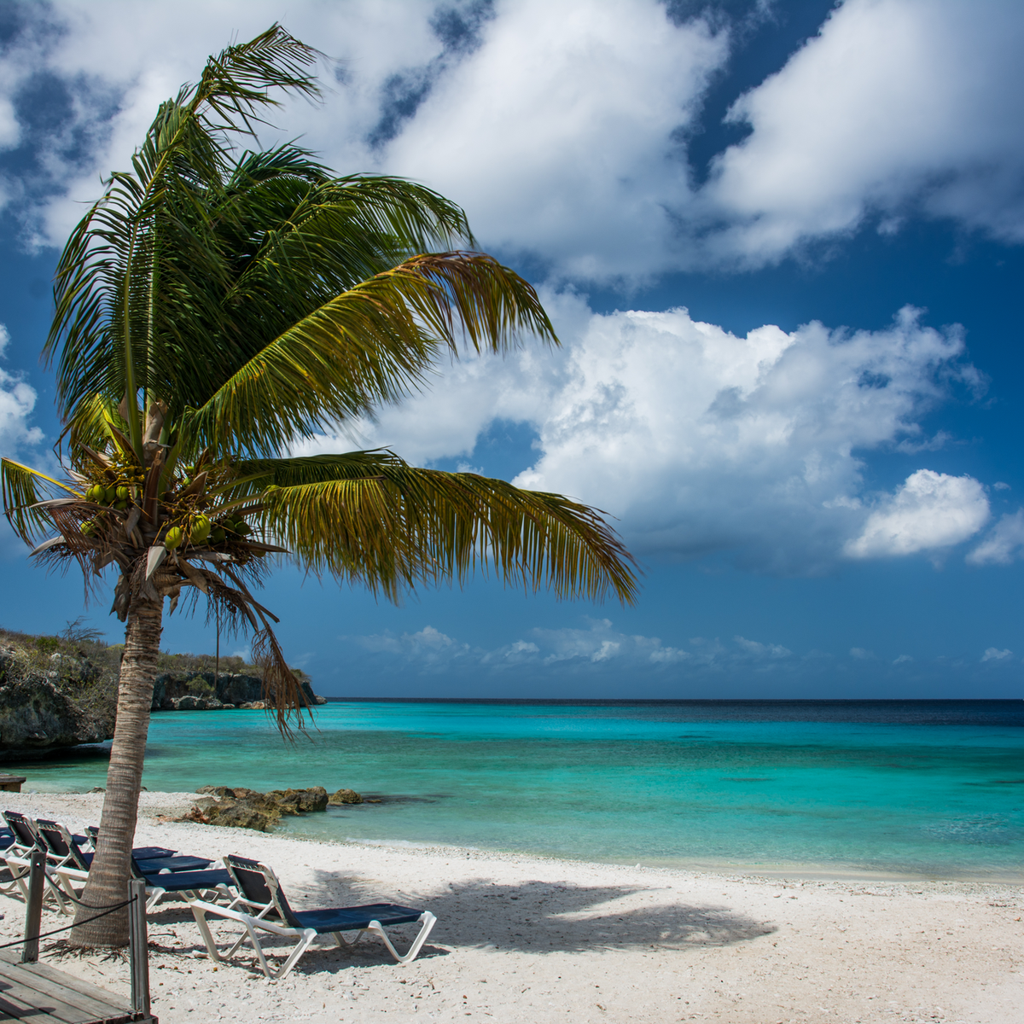} 
        {bpp}
    \end{minipage}\hfill
    \begin{minipage}{0.198\textwidth}
        \centering
        \textbf{PICS~\cite{lei2023text}}
        \includegraphics[width=\textwidth]{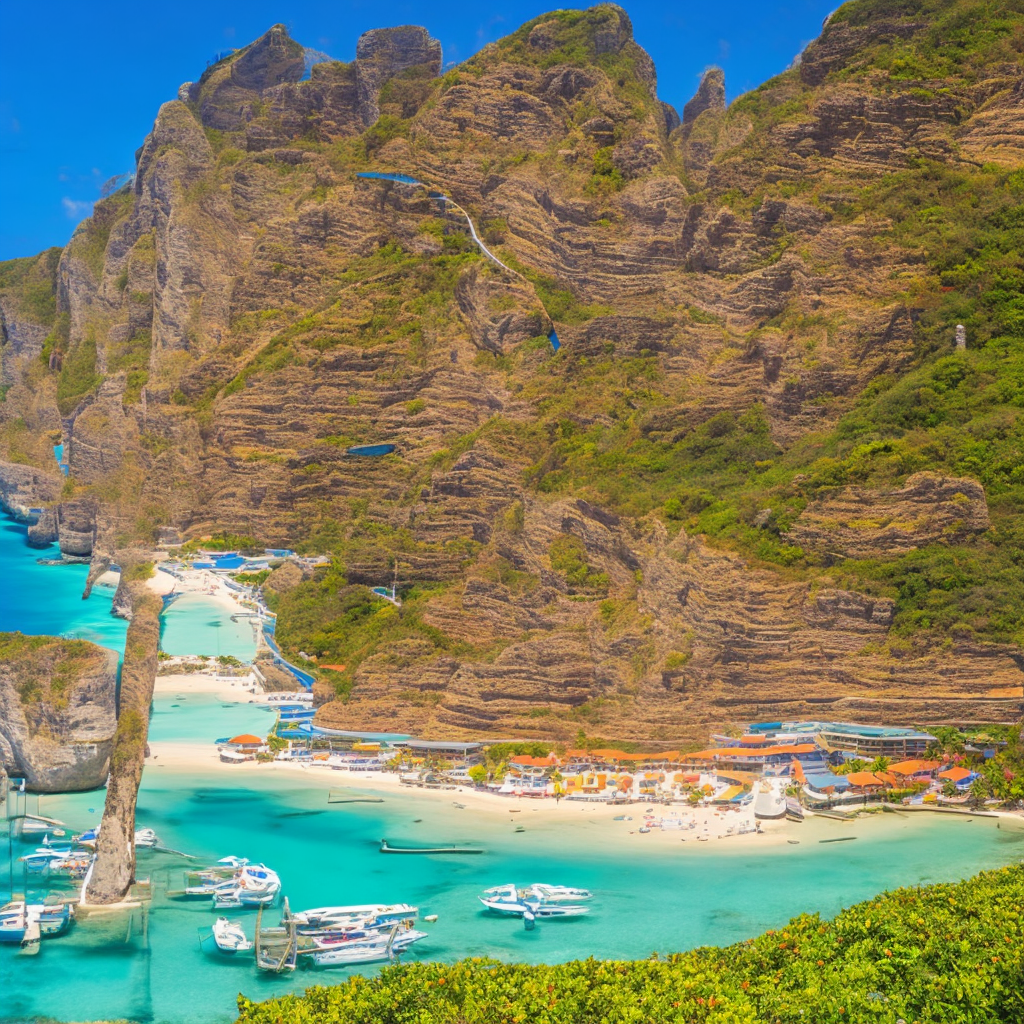}
        {0.01217~\textcolor{lightcoral}{(4.11$\times$)}}
    \end{minipage}\hfill
    \begin{minipage}{0.198\textwidth}
        \centering
        \textbf{MS-ILLM~\cite{msillm}}
        \includegraphics[width=\textwidth]{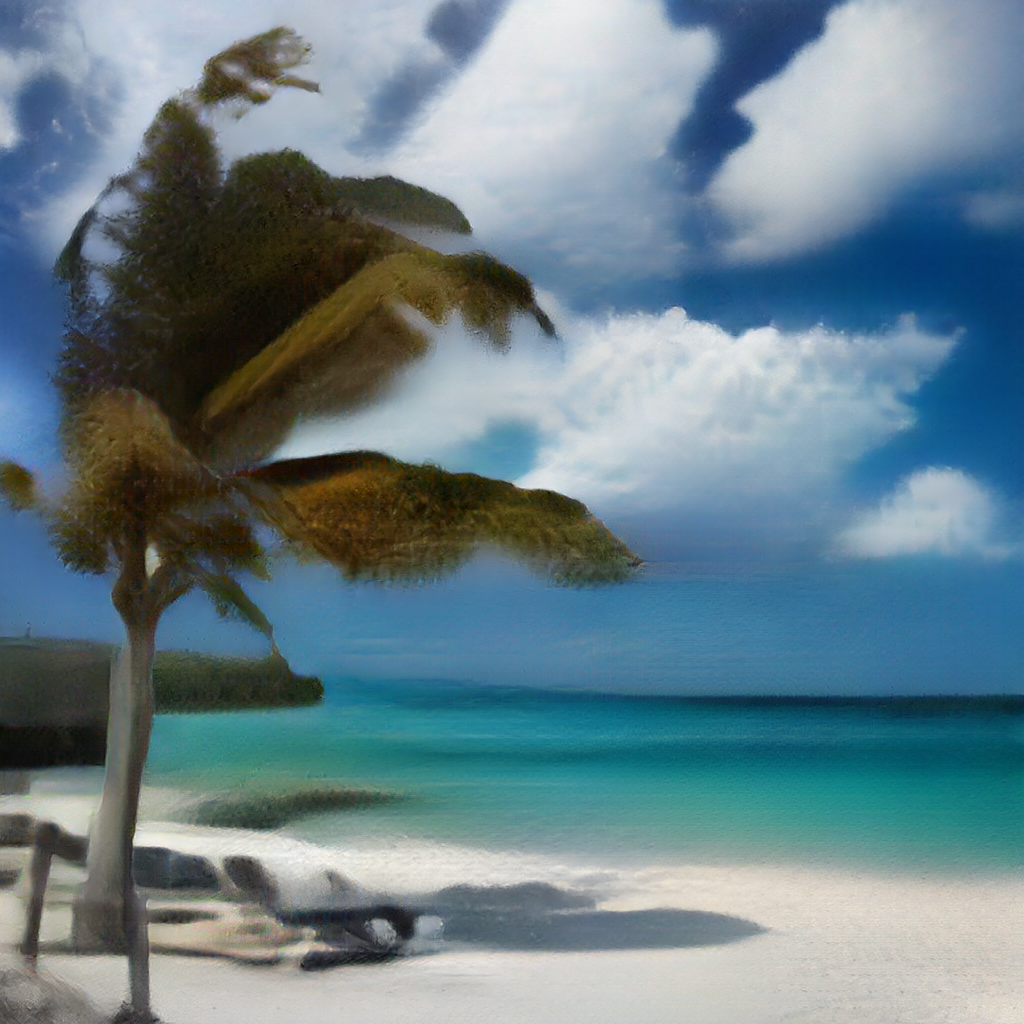}
        {0.00503~\textcolor{lightcoral}{(1.70$\times$)}}
    \end{minipage}\hfill
    \begin{minipage}{0.198\textwidth}
        \centering
        \textbf{PerCo~\cite{perco}}
        \includegraphics[width=\textwidth]{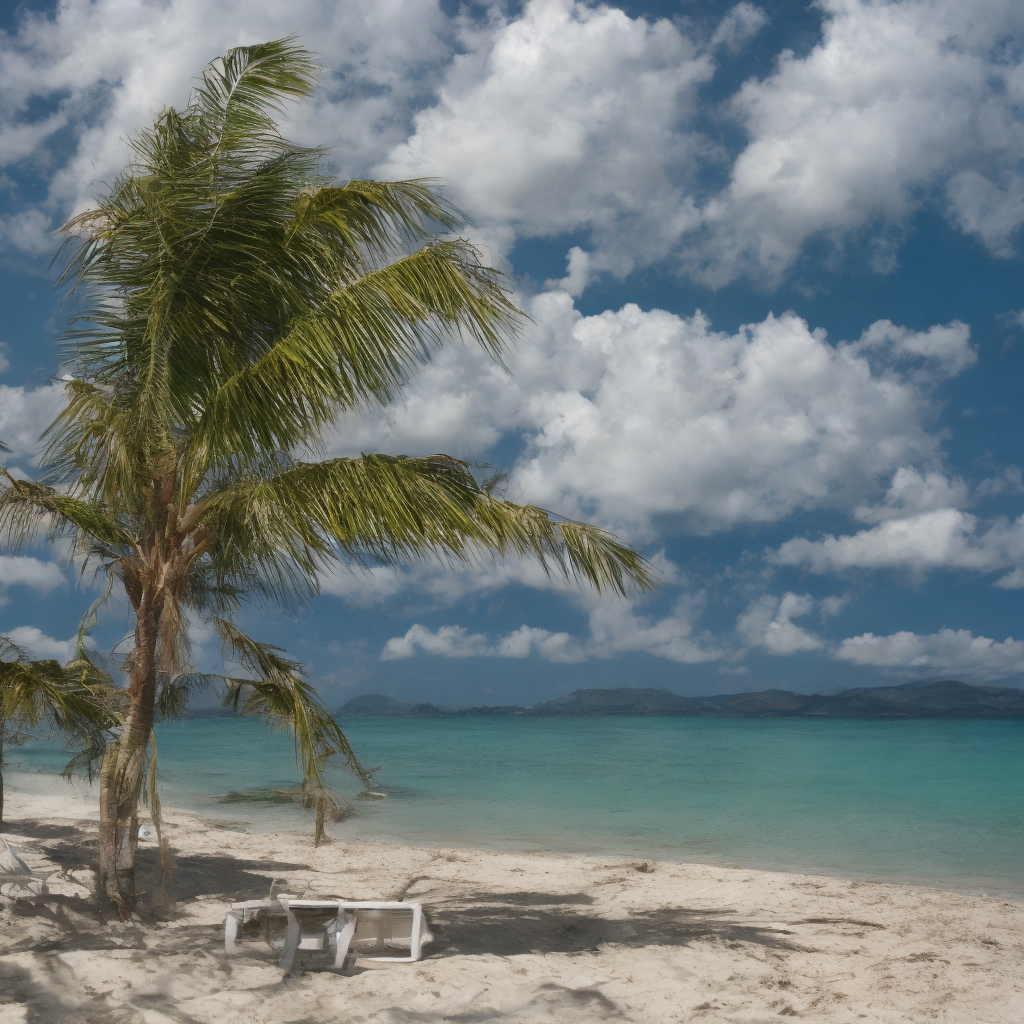}
        {0.00238~\textcolor{lightcoral}{(0.80$\times$)}}
    \end{minipage}\hfill
    \begin{minipage}{0.198\textwidth}
        \centering
        \textbf{Ours (text+image)}
        \includegraphics[width=\textwidth]{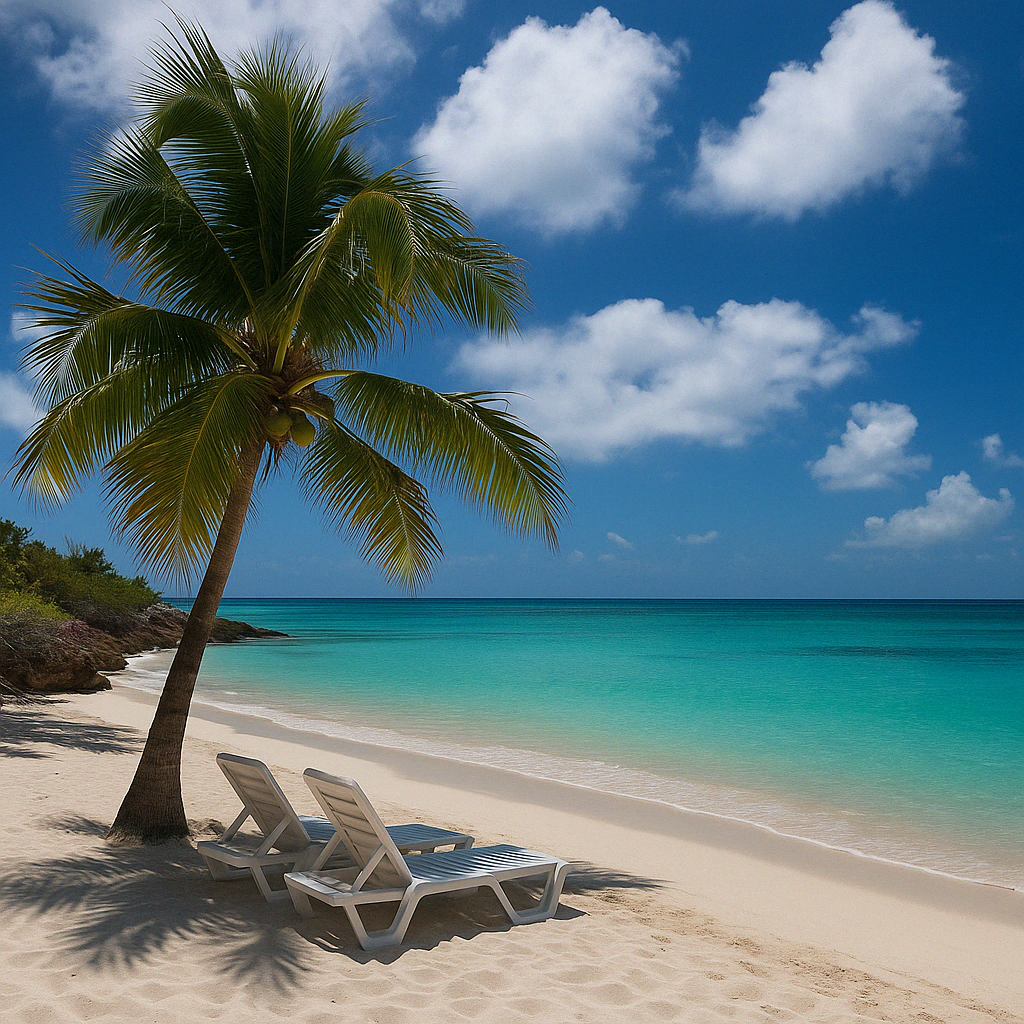}
        {0.00296~\textcolor{lightcoral}{(1.00$\times$)}}
    \end{minipage}


    \begin{minipage}{0.198\textwidth} 
        \centering
        \includegraphics[width=\textwidth]{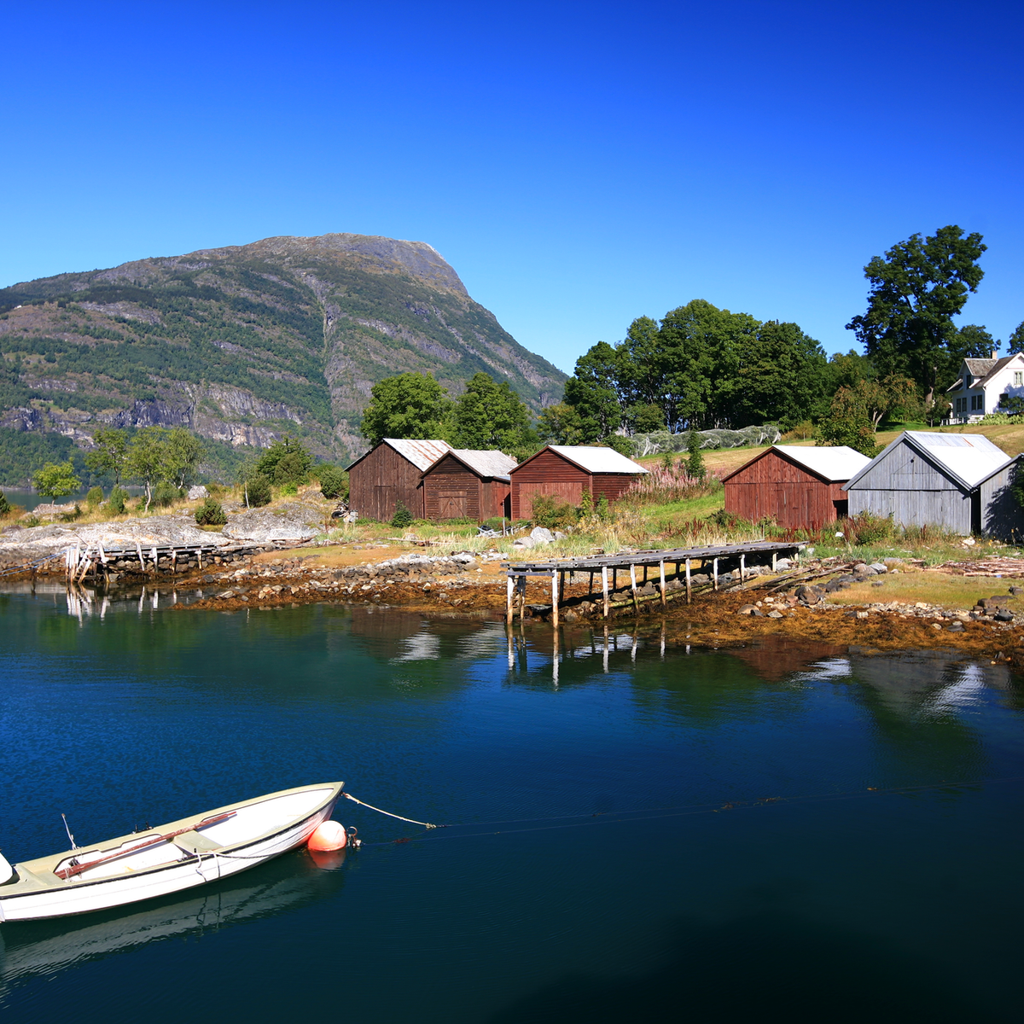} 
        {bpp}
    \end{minipage}\hfill
    \begin{minipage}{0.198\textwidth}
        \centering
        \includegraphics[width=\textwidth]{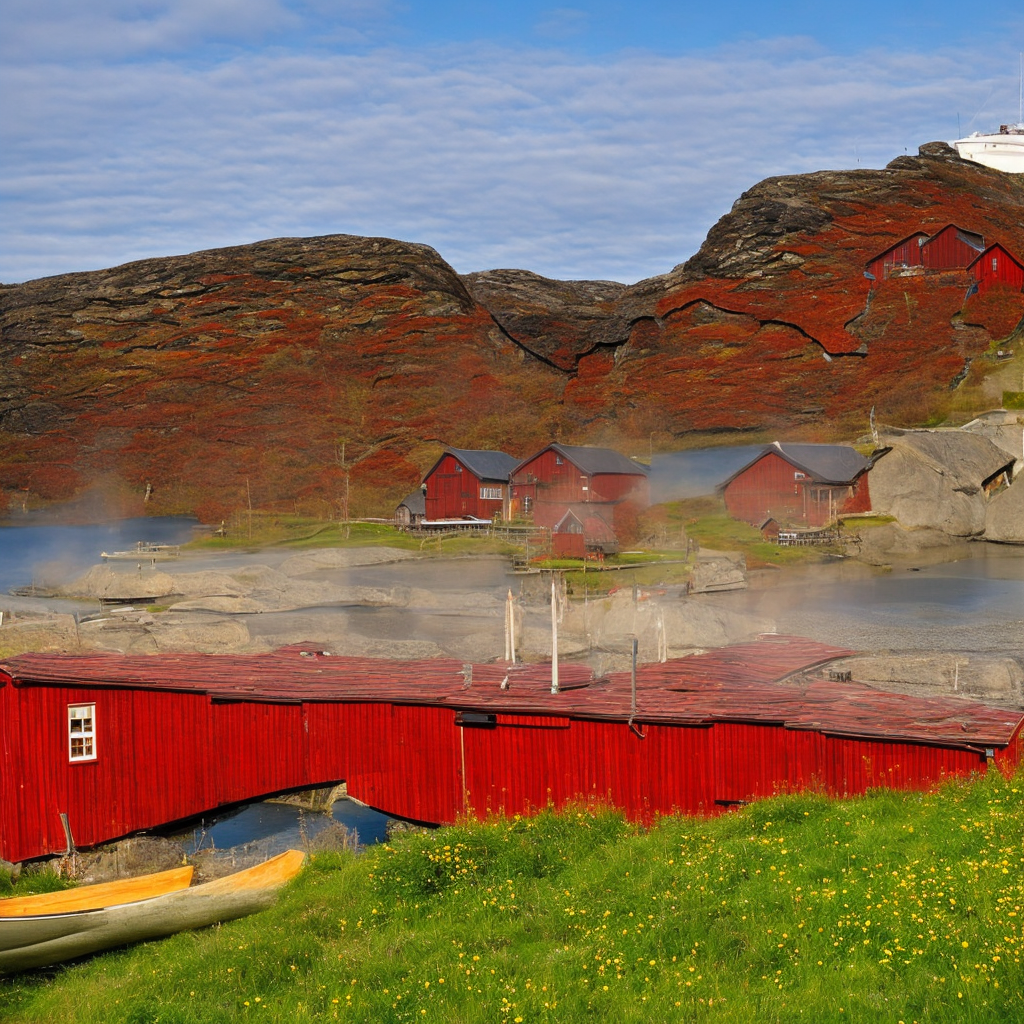}
        {0.01001~\textcolor{lightcoral}{(2.13$\times$)}}
    \end{minipage}\hfill
    \begin{minipage}{0.198\textwidth}
        \centering
        \includegraphics[width=\textwidth]{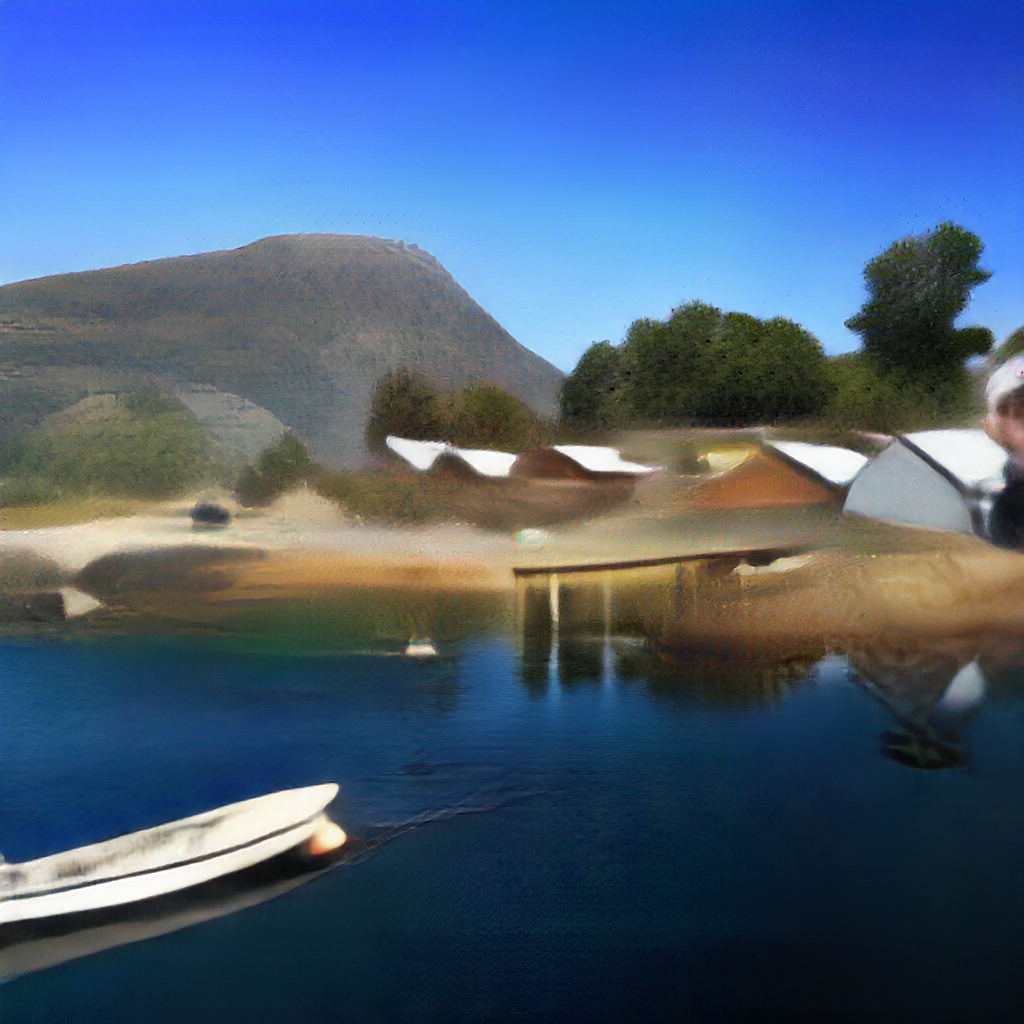}
        {0.00530~\textcolor{lightcoral}{(1.13$\times$)}}
    \end{minipage}\hfill
    \begin{minipage}{0.198\textwidth}
        \centering
        \includegraphics[width=\textwidth]{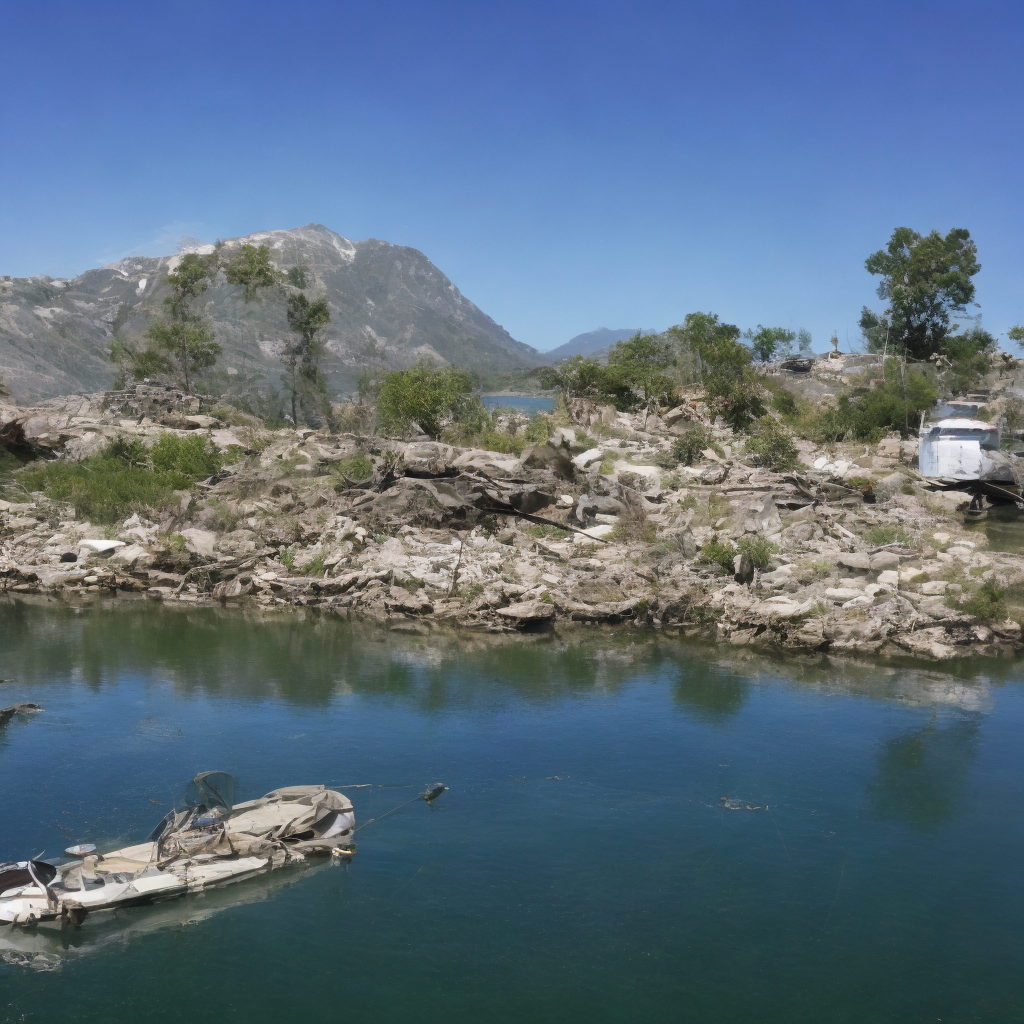}
        {0.00227~\textcolor{lightcoral}{(0.48$\times$)}}
    \end{minipage}\hfill
    \begin{minipage}{0.198\textwidth}
        \centering
        \includegraphics[width=\textwidth]{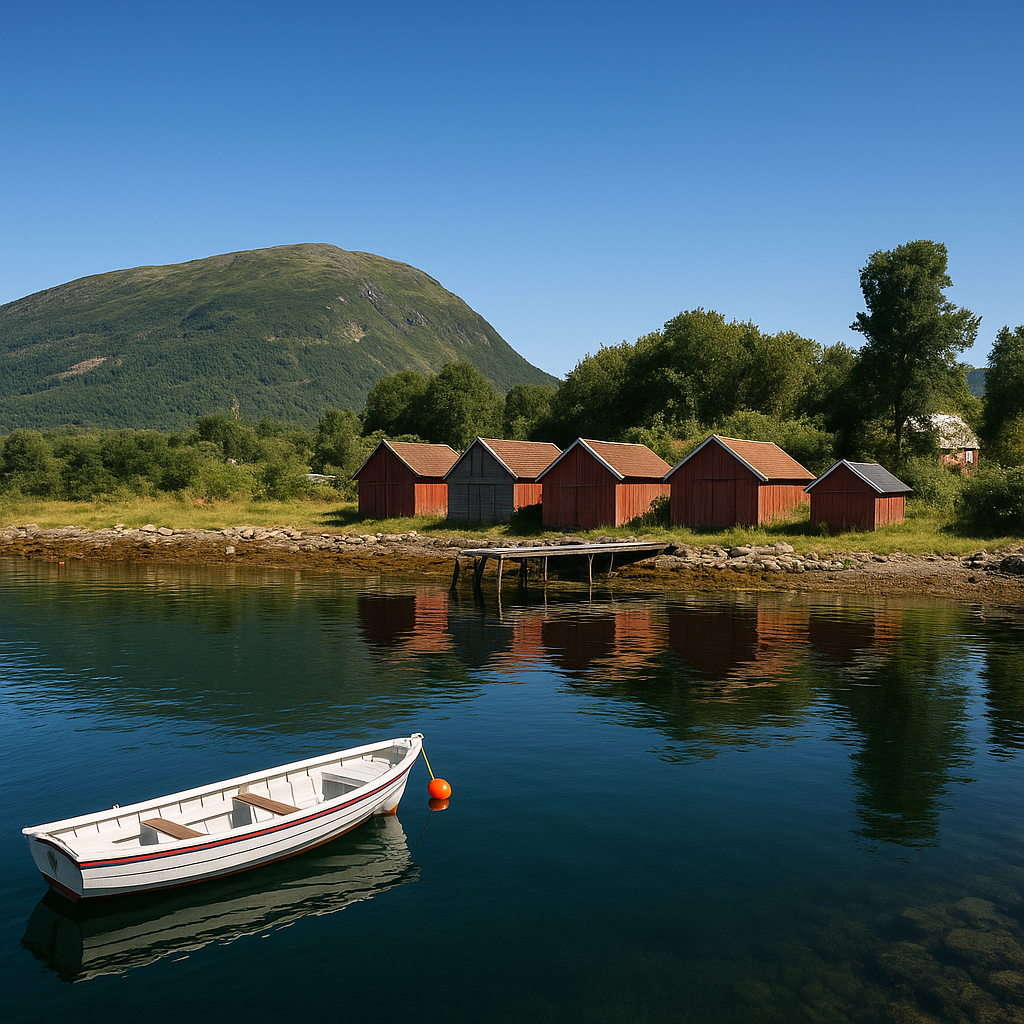}
        {0.00470~\textcolor{lightcoral}{(1.00$\times$)}}
    \end{minipage}

    \begin{minipage}{0.198\textwidth} 
        \centering
        \includegraphics[width=\textwidth]{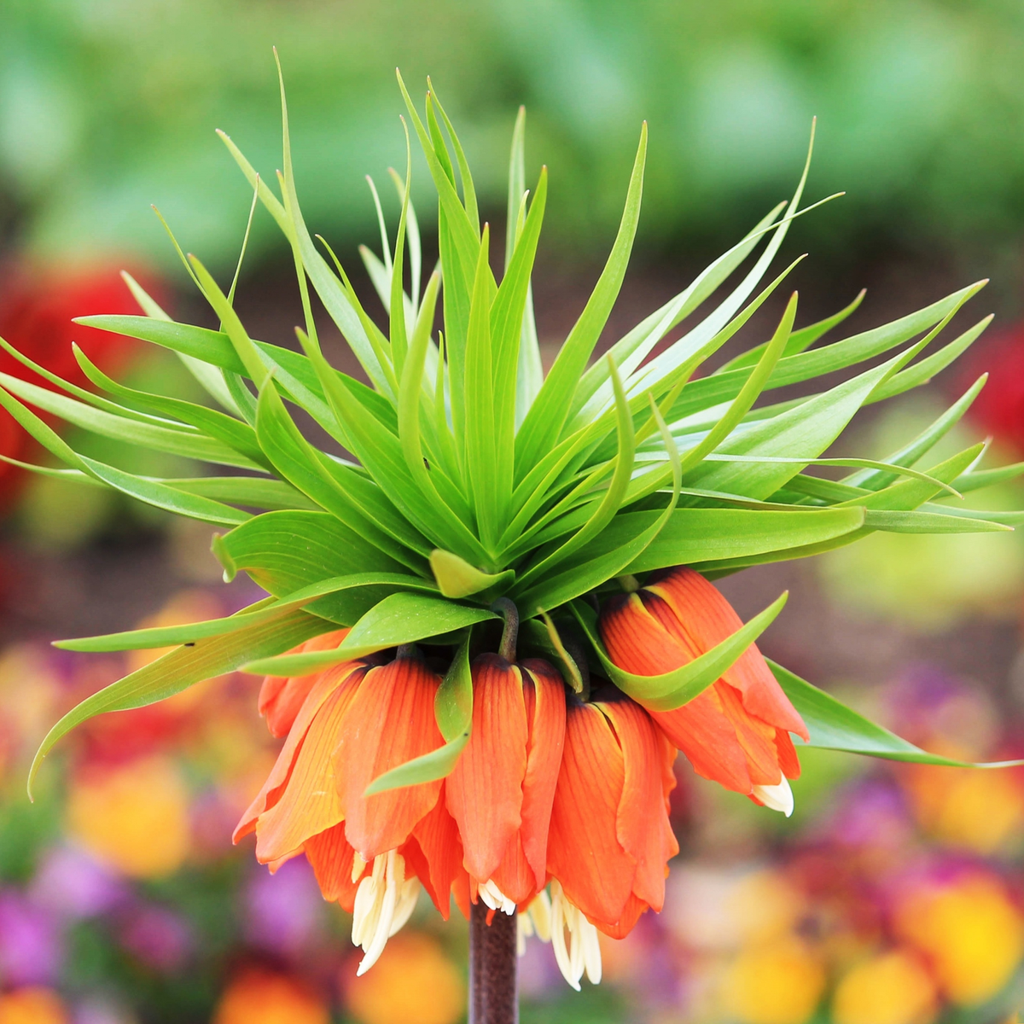} 
        {bpp}
    \end{minipage}\hfill
    \begin{minipage}{0.198\textwidth}
        \centering
        \includegraphics[width=\textwidth]{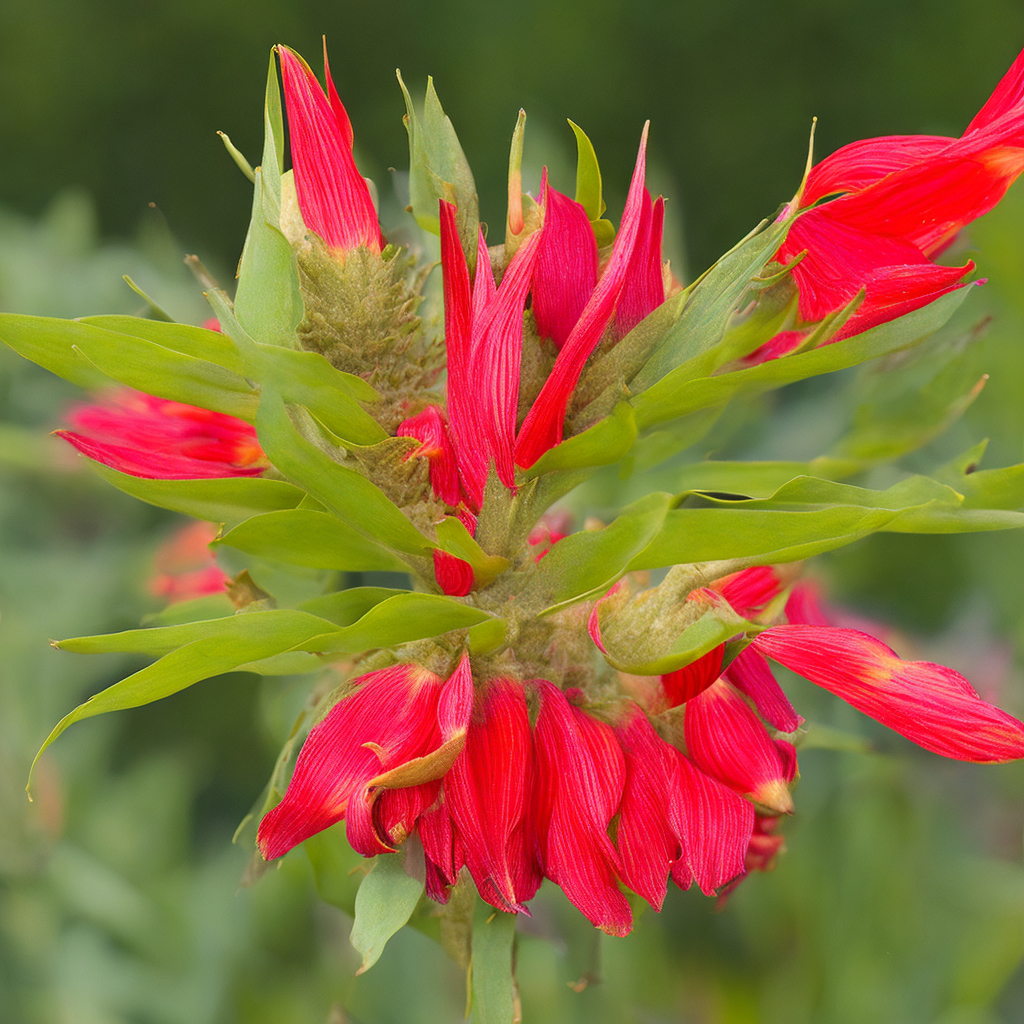}
        {0.01459~\textcolor{lightcoral}{(2.43$\times$)}}
    \end{minipage}\hfill
    \begin{minipage}{0.198\textwidth}
        \centering
        \includegraphics[width=\textwidth]{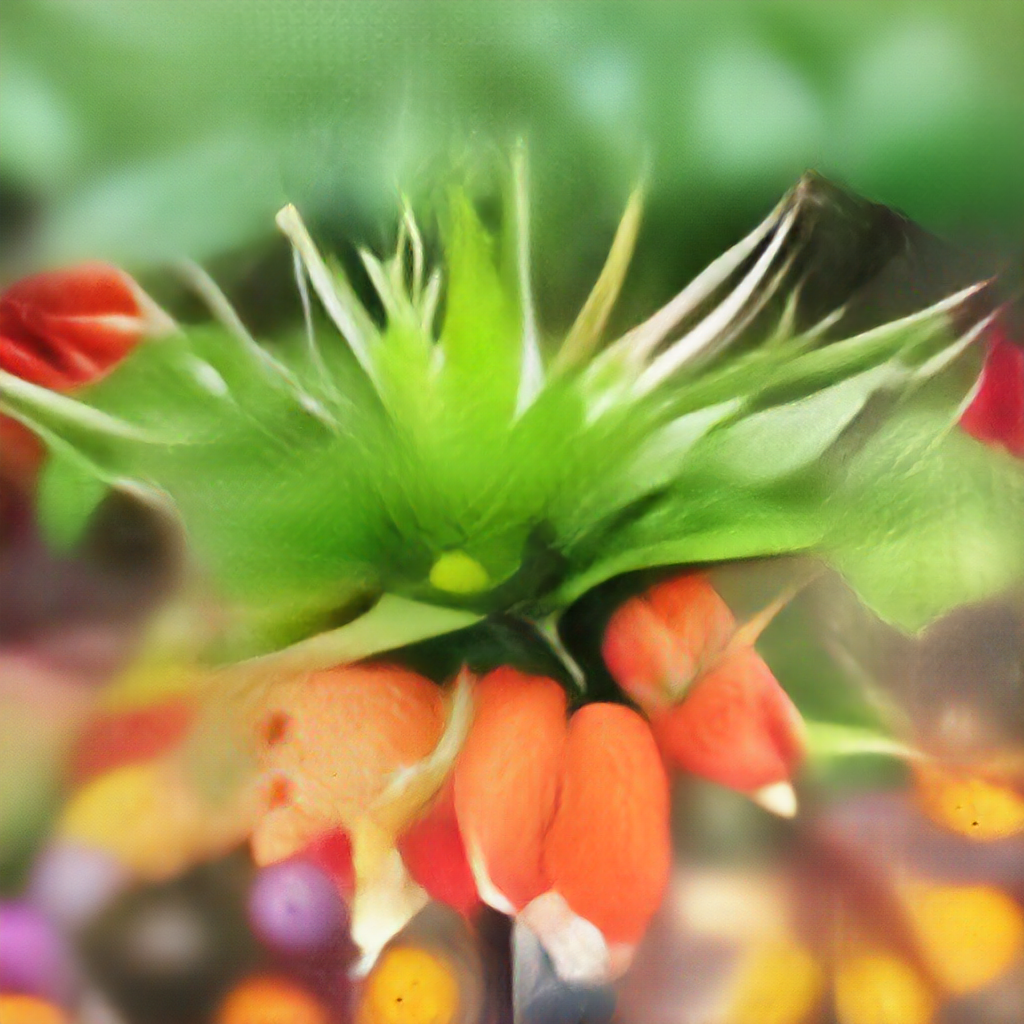}
        {0.00633~\textcolor{lightcoral}{(1.06$\times$)}}
    \end{minipage}\hfill
    \begin{minipage}{0.198\textwidth}
        \centering
        \includegraphics[width=\textwidth]{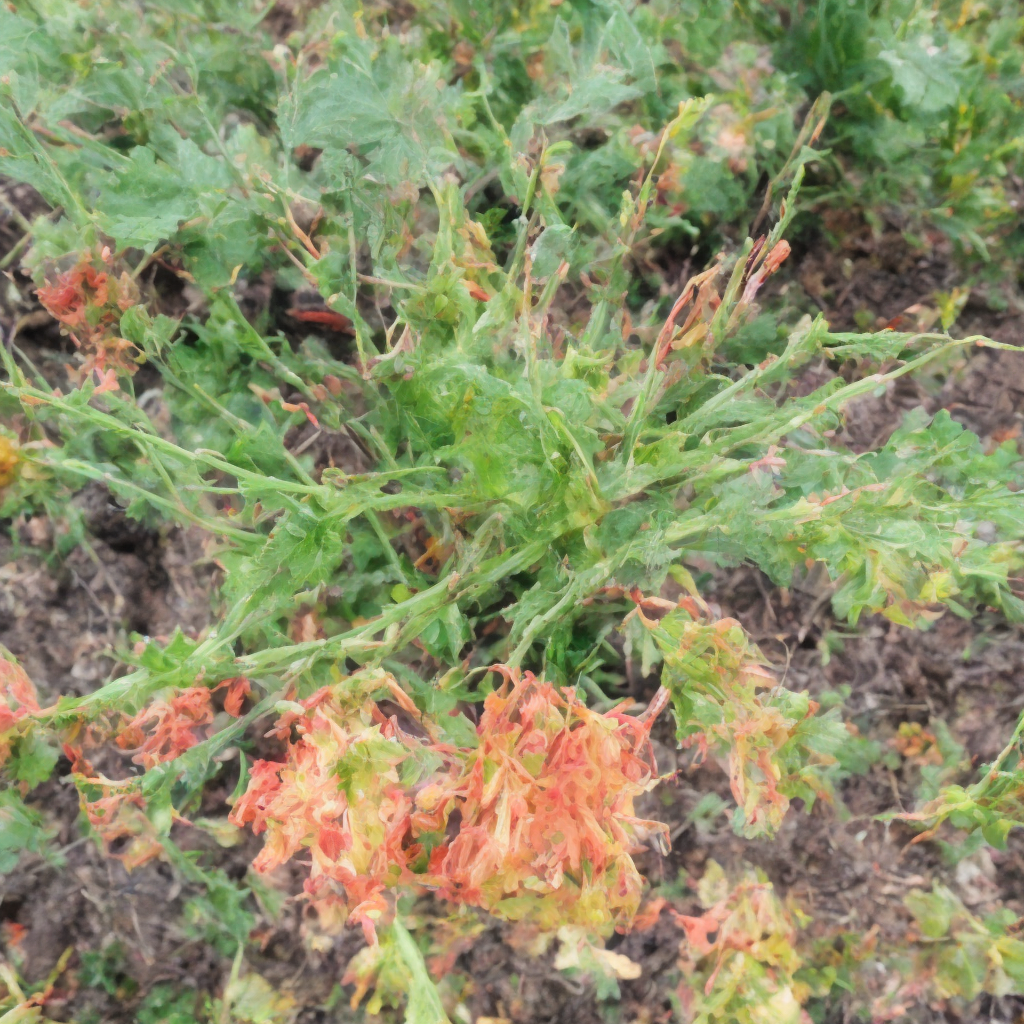}
        {0.00235~\textcolor{lightcoral}{(0.39$\times$)}}
    \end{minipage}\hfill
    \begin{minipage}{0.198\textwidth}
        \centering
        \includegraphics[width=\textwidth]{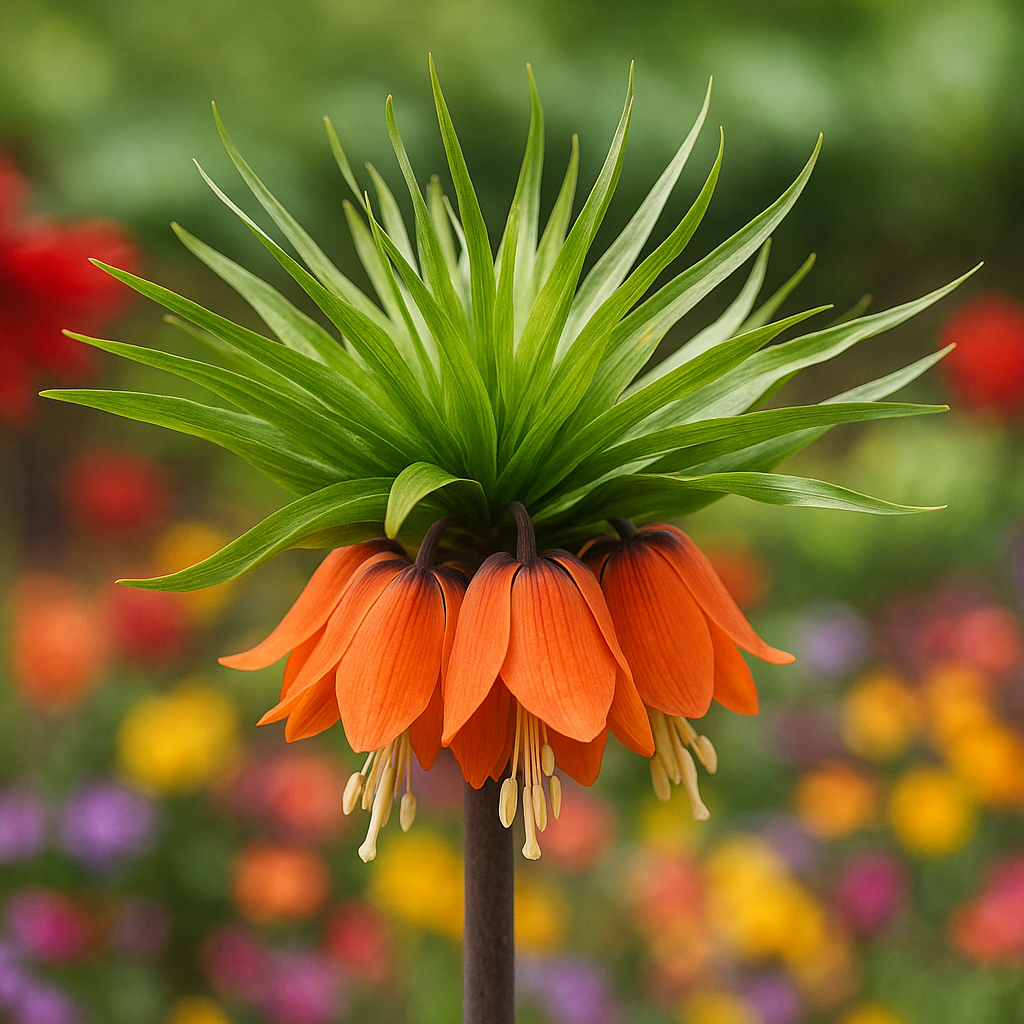}
        {0.00599~\textcolor{lightcoral}{(1.00$\times$)}}
    \end{minipage}
    
    \caption{The comparison of qualitative results between the baselines and ours (text+image) method is provided, with the bits per pixel (bpp) for each image listed below. The \textcolor{lightcoral}{rate multiples} comparing other methods with ours are provided in parentheses.}
    \label{fig:qualitative_results_t+i}
\end{figure*}

\myparagraph{Baselines}\quad
We include three representative perceptual codecs capable of achieving ultra-low bitrate image compression for comparison: MS-ILLM~\cite{msillm}, Text + Sketch~\cite{lei2023text}, and PerCo~\cite{perco}. 
Among them, MS-ILLM is a state-of-the-art perceptual image codec that leverages adversarial training to enhance the realism of reconstructed images. 
Text+Sketch utilizes pretrained text-to-image generative models~\cite{latentdiffusion} to reconstruct images from compressed text descriptions and sketches, achieving high-fidelity reconstructions at ultra-low bitrates without end-to-end training. 
PerCo employs a diffusion-based decoder conditioned on vector-quantized image representations and global image descriptions, enabling realistic image reconstructions at bitrates as low as 0.003 bits per pixel. For details on replication of baselines, please see App ~\ref{sec:app_expermential_details}.

\subsection{Main Results}
\myparagraph{Quantitative Results}\quad
In Figure~\ref{fig:main_results}, we compare our results to state-of-the-art codecs on both perceptual quality metrics and consistency metrics. Across all perceptual metrics, our method consistently achieves higher scores than existing approaches at similar or even lower bitrates, demonstrating its superior visual quality at ultra-low bitrates. 
In terms of consistency metrics, our method achieves the best performance in CLIPSIM, indicating stronger preservation of semantic content. 
Additionally, we observe our approach yields significantly lower (better) results in DISTS compared to Text + Sketch and MS-ILLM. 
Although it does not surpass PerCo in DISTS, it shows clear advantages in the other three metrics, highlighting its ability to balance aesthetic quality and semantic consistency at ultra-low bitrates. Notably, our method requires no additional training, making it more accessible and versatile.

These results highlight the ability of our method to balance aesthetic quality and semantic consistency under ultra-low bitrates, offering a strong alternative to existing perceptual coding techniques. At the same time, they also reveal the potential of GPT-4o image generation in perceptual image coding at ultra-low bitrate.



\myparagraph{Qualitative Results}\quad
Fig.~\ref{fig:qualitative_results_t+i} presents a qualitative comparison of partial decoded images. Below each image, the corresponding bitrate is listed. From the subjective results, it can be observed that some of the comparison methods exhibit noticeable structural distortions, leading to inaccurate semantic transmission, while others suffer from errors or blurring in texture details, severely affecting the image quality. In contrast, thanks to the powerful generative capabilities of GPT-4o, our approach maintains a better balance between semantic and structural fidelity while still producing high-quality decoded images, achieving a superior trade-off between perceptual quality and image consistency.

\begin{figure}[htbp]
    \centering

    \begin{minipage}{0.157\textwidth} 
        \centering
        \textbf{Origin}
        \includegraphics[width=\textwidth]{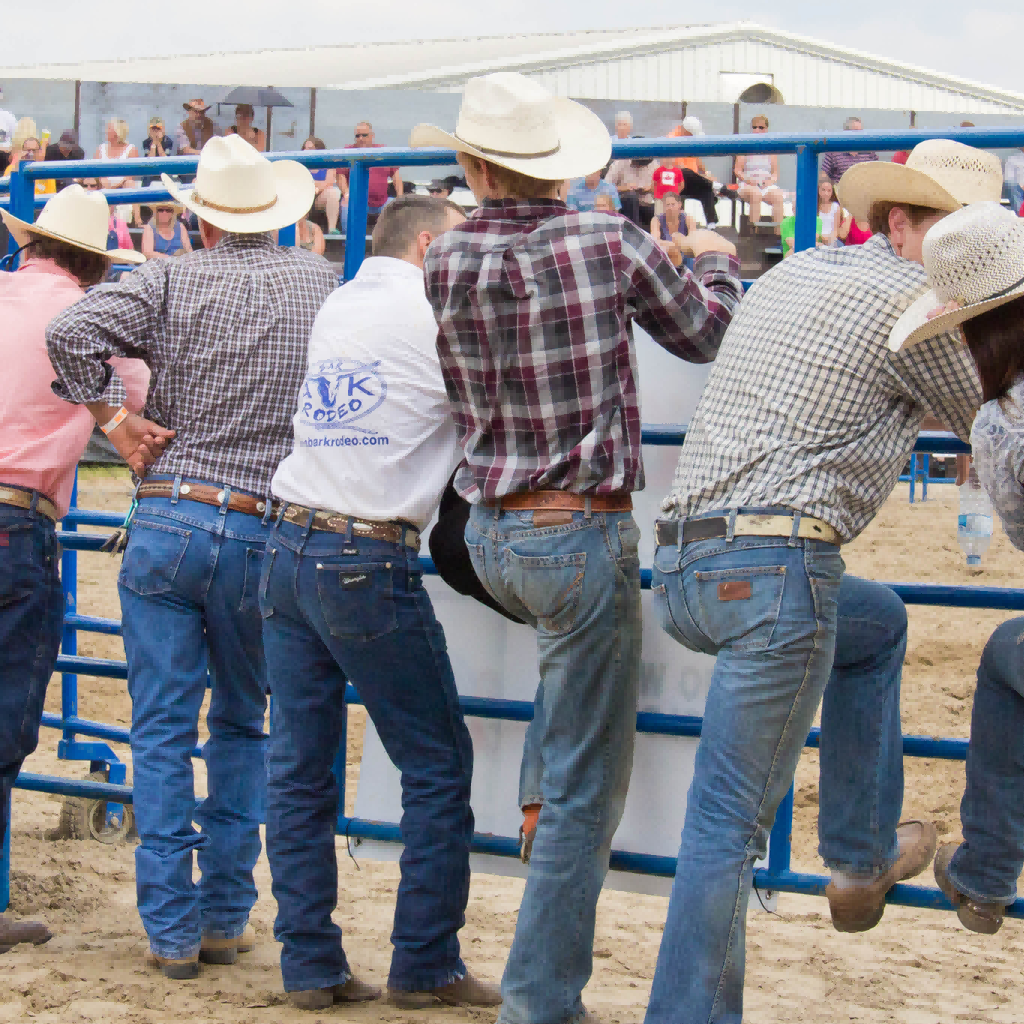} 
        {bpp}
    \end{minipage}\hfill
    \begin{minipage}{0.157\textwidth}
        \centering
        \textbf{PIC~\cite{lei2023text}}
        \includegraphics[width=\textwidth]{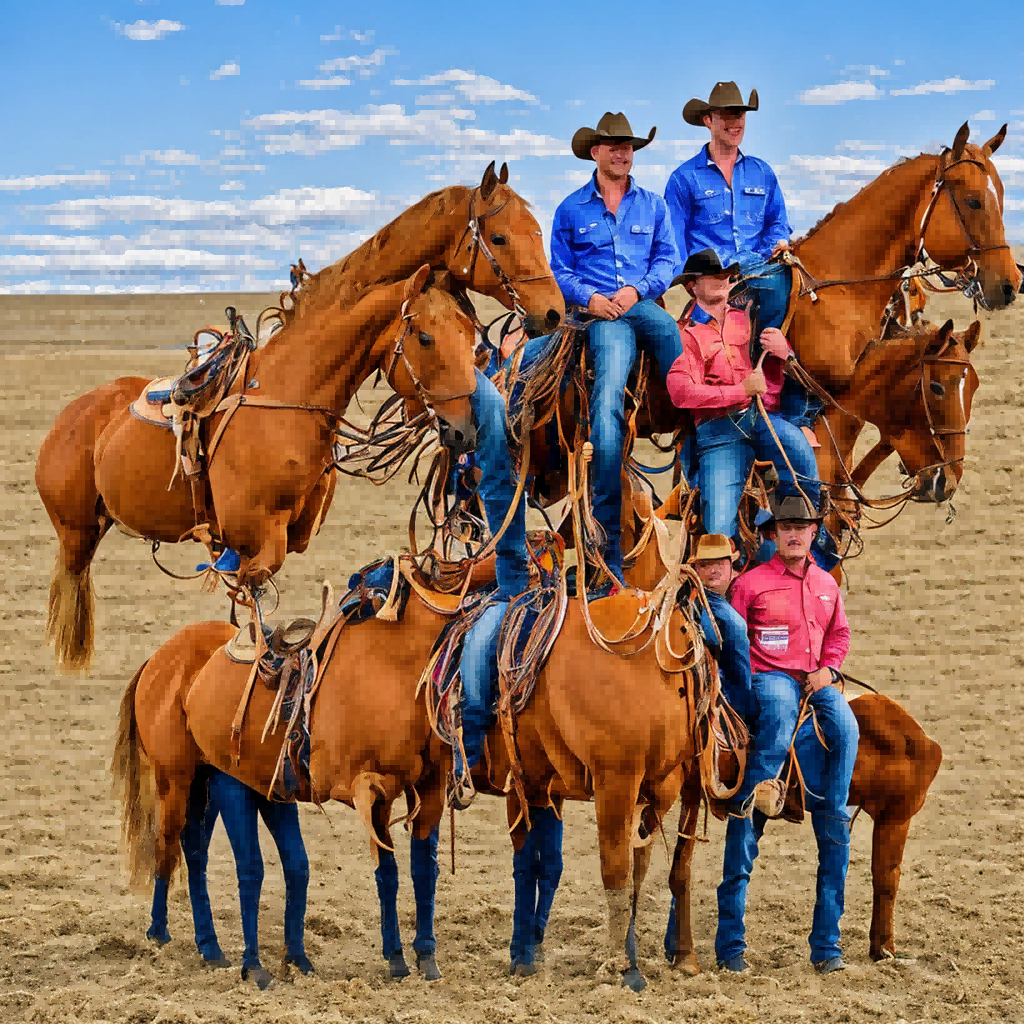}
        {0.00106}
    \end{minipage}\hfill
    \begin{minipage}{0.157\textwidth}
        \centering
        \textbf{Ours (text)}
        \includegraphics[width=\textwidth]{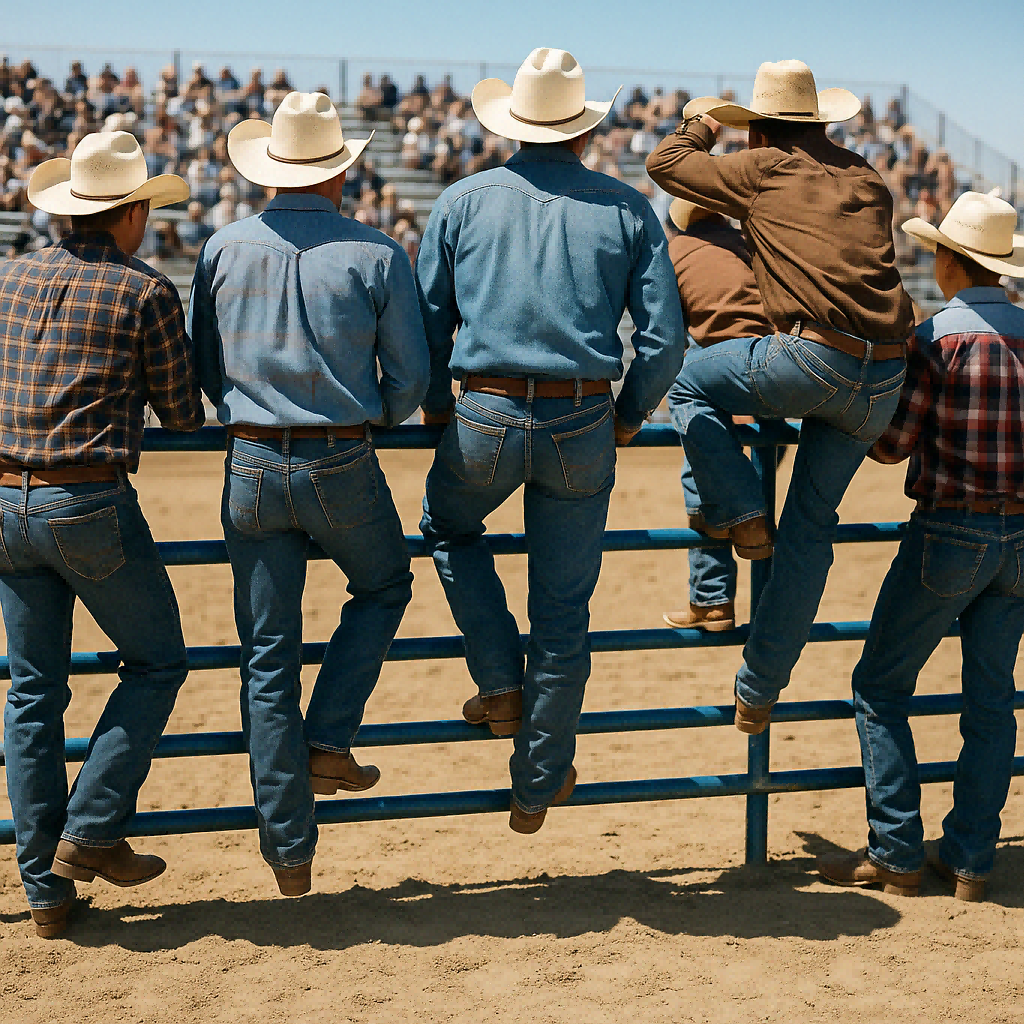}
        {0.00175}
    \end{minipage}

    \begin{minipage}{0.157\textwidth} 
        \centering
        \includegraphics[width=\textwidth]{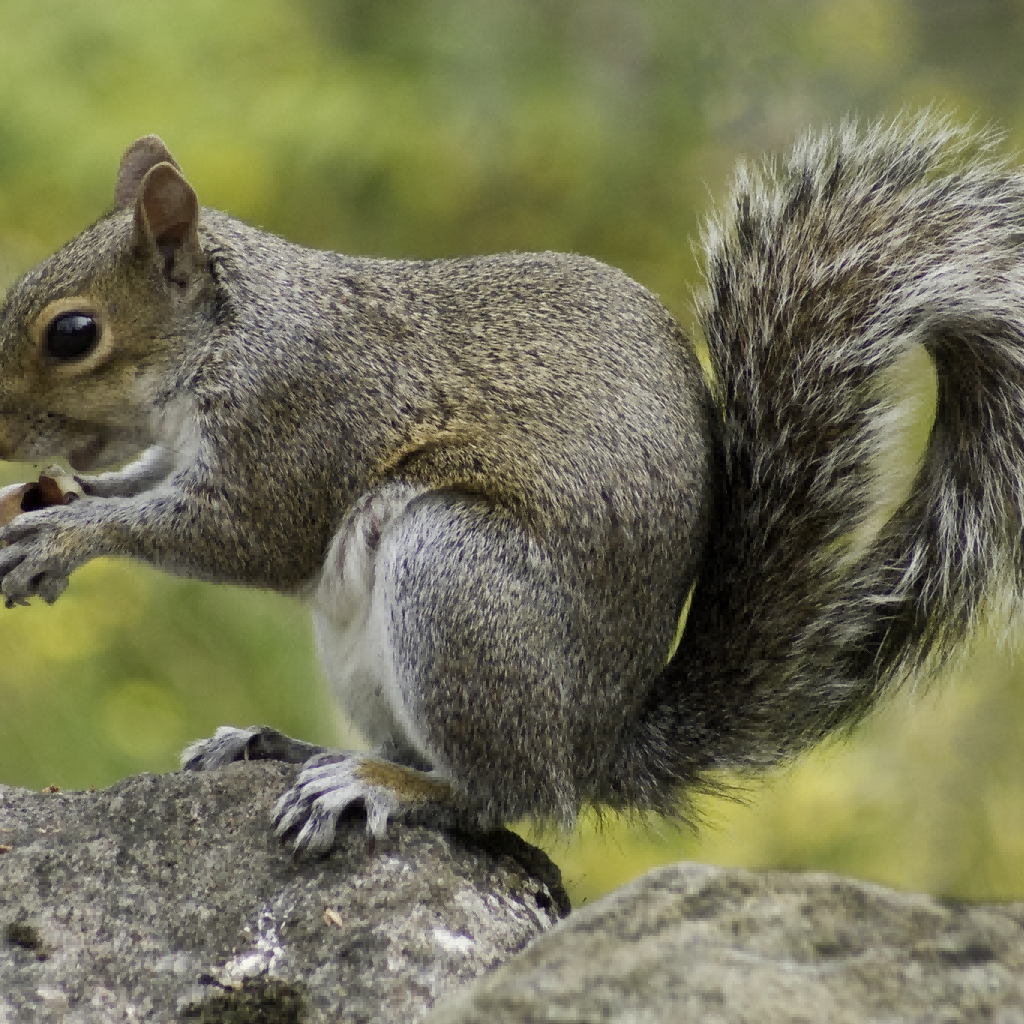} 
        {bpp}
    \end{minipage}\hfill
    \begin{minipage}{0.157\textwidth}
        \centering
        \includegraphics[width=\textwidth]{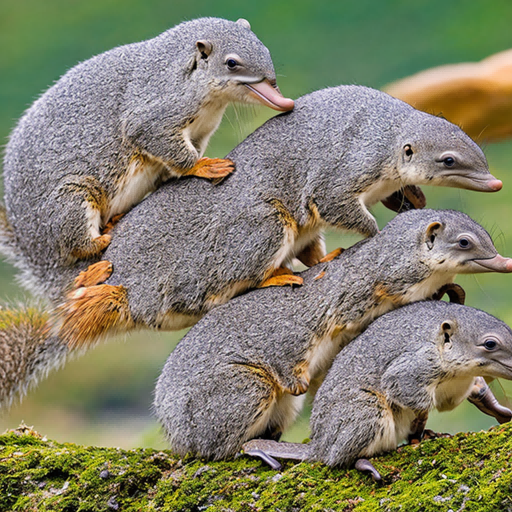}
        {0.00106}
    \end{minipage}\hfill
    \begin{minipage}{0.157\textwidth}
        \centering
        \includegraphics[width=\textwidth]{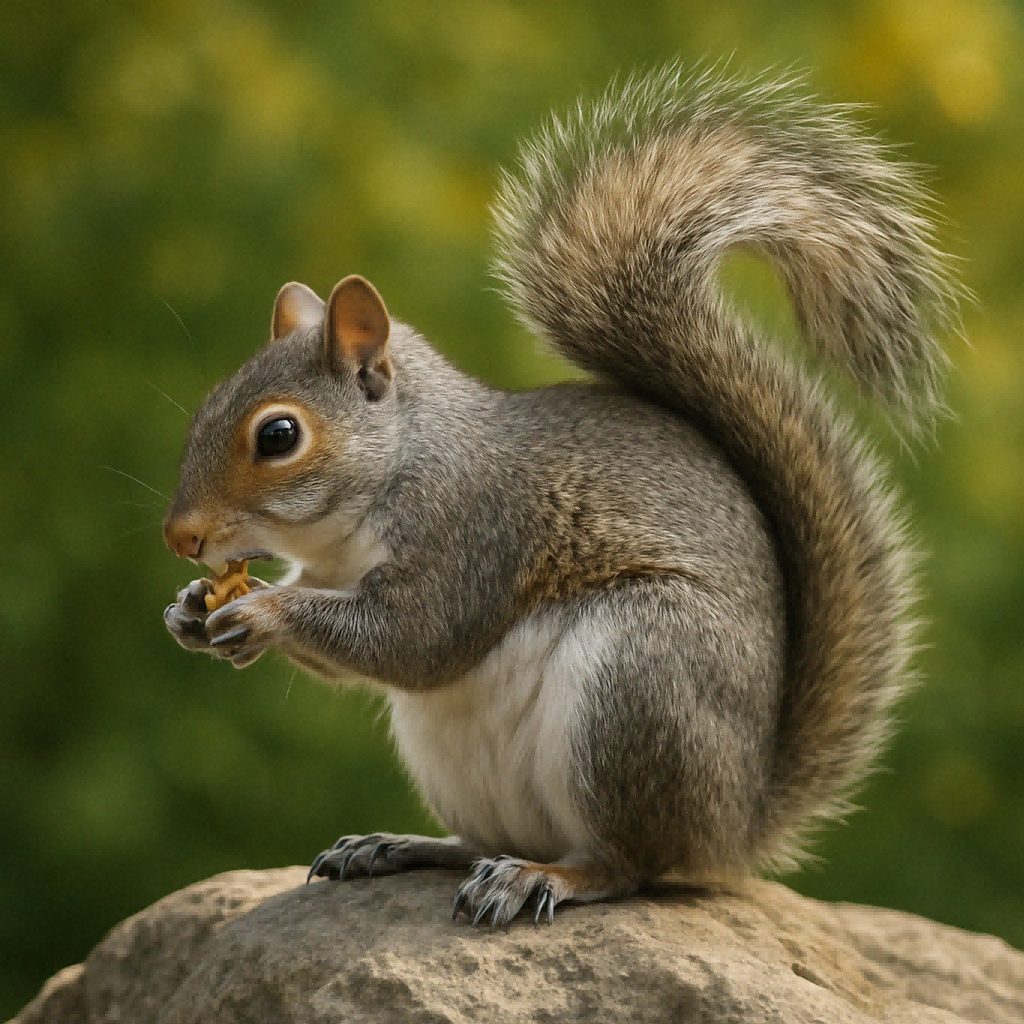}
        {0.00183}
    \end{minipage}
    
    \caption{Comparison of results generated using only text.}
    \label{fig:qualitative_results_t}
\end{figure}

We also explored the semantic coding capability of the framework based on pure test transmission, with comparison results shown in Fig.~\ref{fig:qualitative_results_t}. Compared to the pure text compression results of Text+Sketch~\cite{lei2023text}, our results are better at preserving the semantic information of the original image.

\begin{figure}[htbp]
    \centering

    \begin{minipage}{0.02\textwidth} 
        \vfill
        \centering
        \rotatebox{90}{\textbf{No Image}} 
        \vfill
    \end{minipage}%
    \hfill
    \begin{minipage}{0.151\textwidth} 
        \centering
        \textbf{No Text}
        \includegraphics[width=\textwidth]{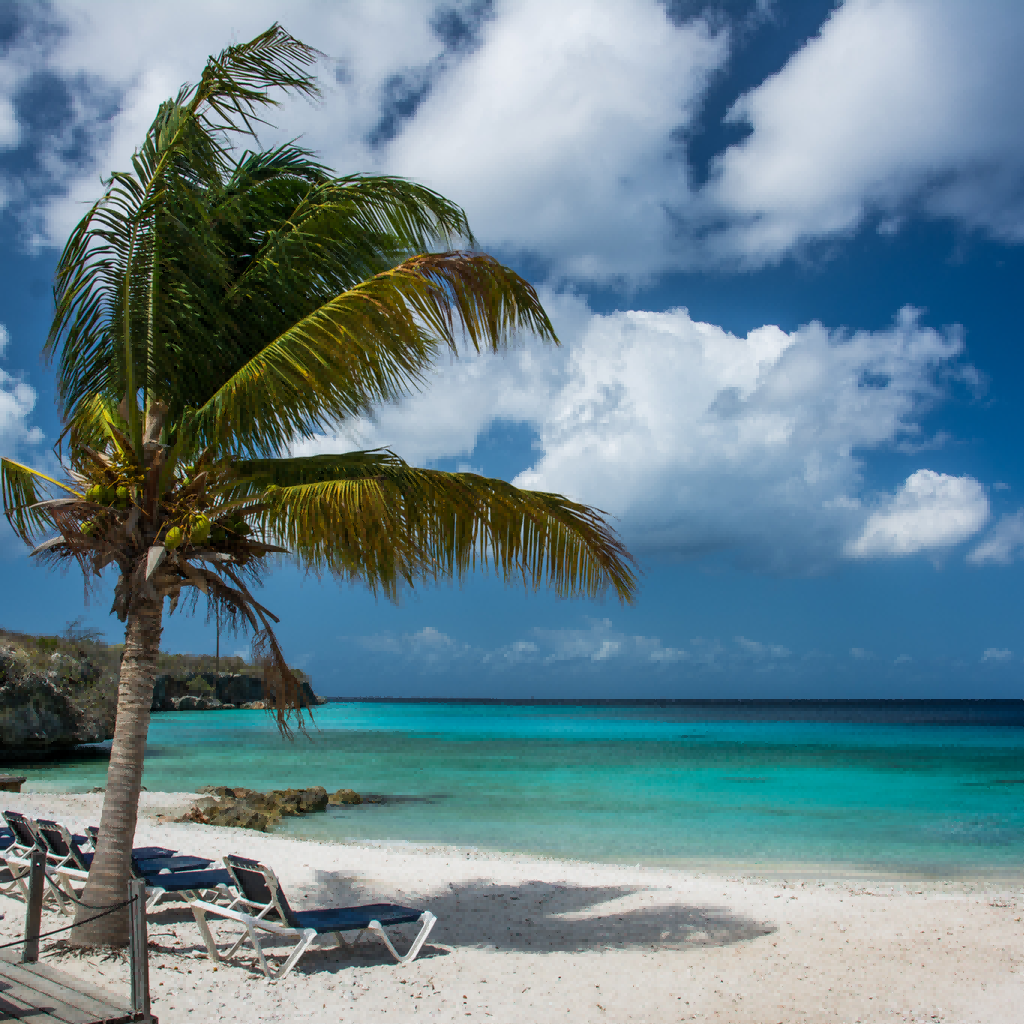}
        {bpp}
    \end{minipage}\hfill
    \begin{minipage}{0.151\textwidth}
        \centering
        \textbf{Short Text}
        \includegraphics[width=\textwidth]{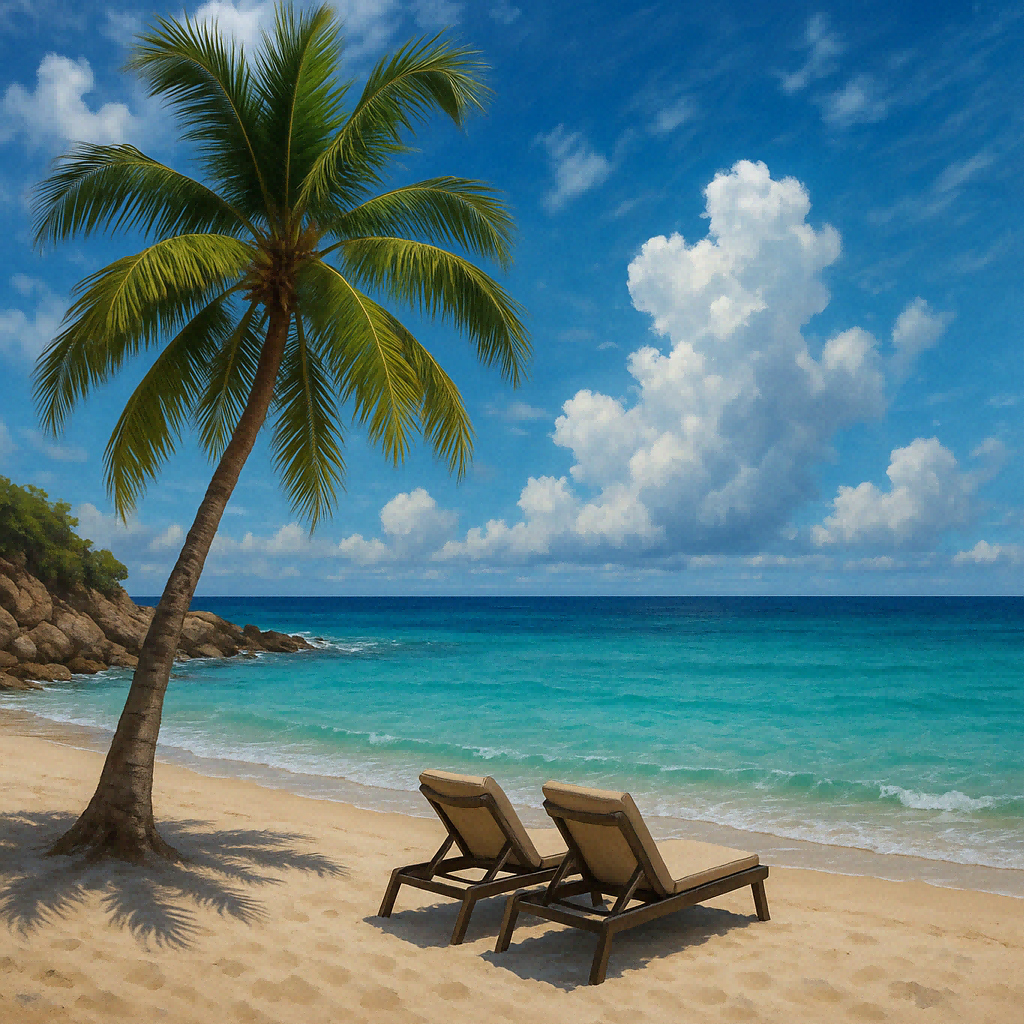}
        {0.00073}
    \end{minipage}\hfill
    \begin{minipage}{0.151\textwidth}
        \centering
        \textbf{Long Text}
        \includegraphics[width=\textwidth]{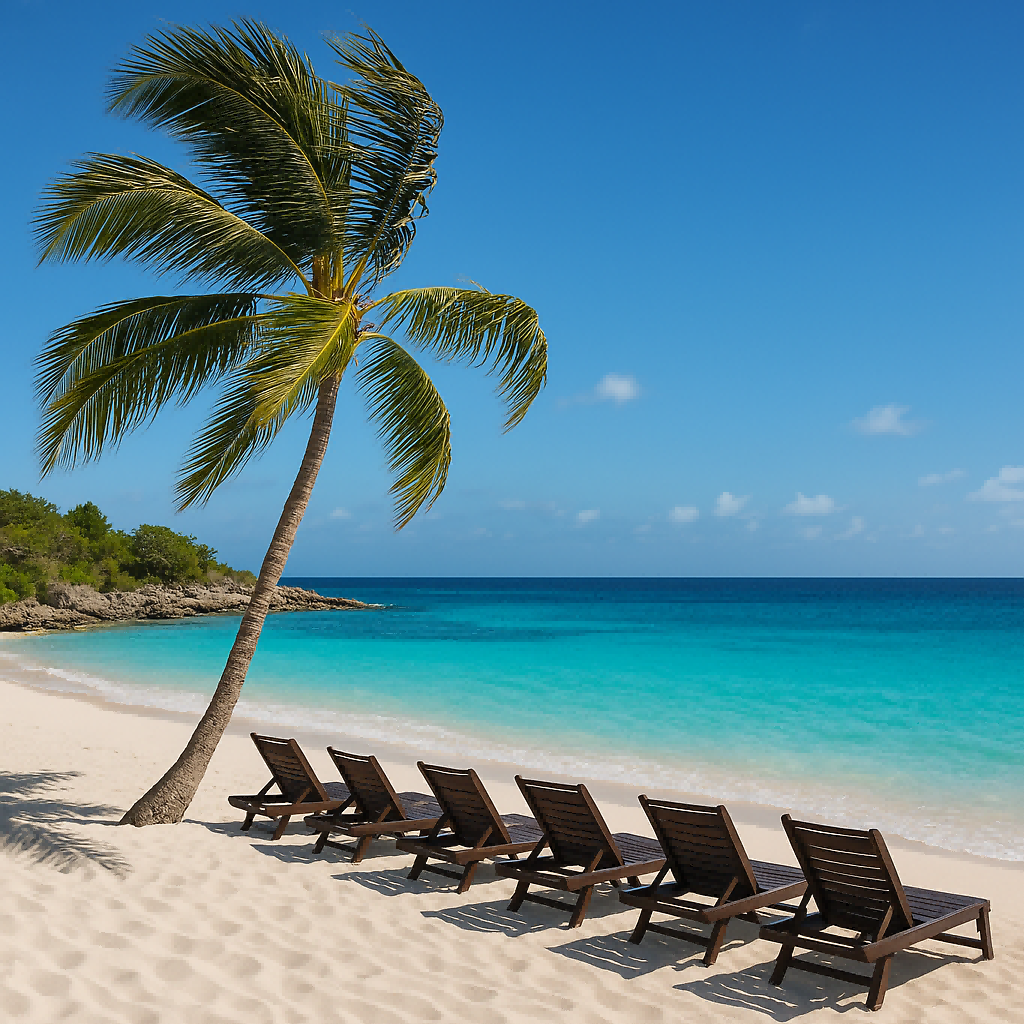}
        {0.00115}
    \end{minipage}

    \begin{minipage}{0.02\textwidth} 
        \vfill
        \centering
        \rotatebox{90}{\textbf{LQ Image}} 
        \vfill
    \end{minipage}%
    \hfill
    \begin{minipage}{0.151\textwidth} 
        \centering
        \includegraphics[width=\textwidth]{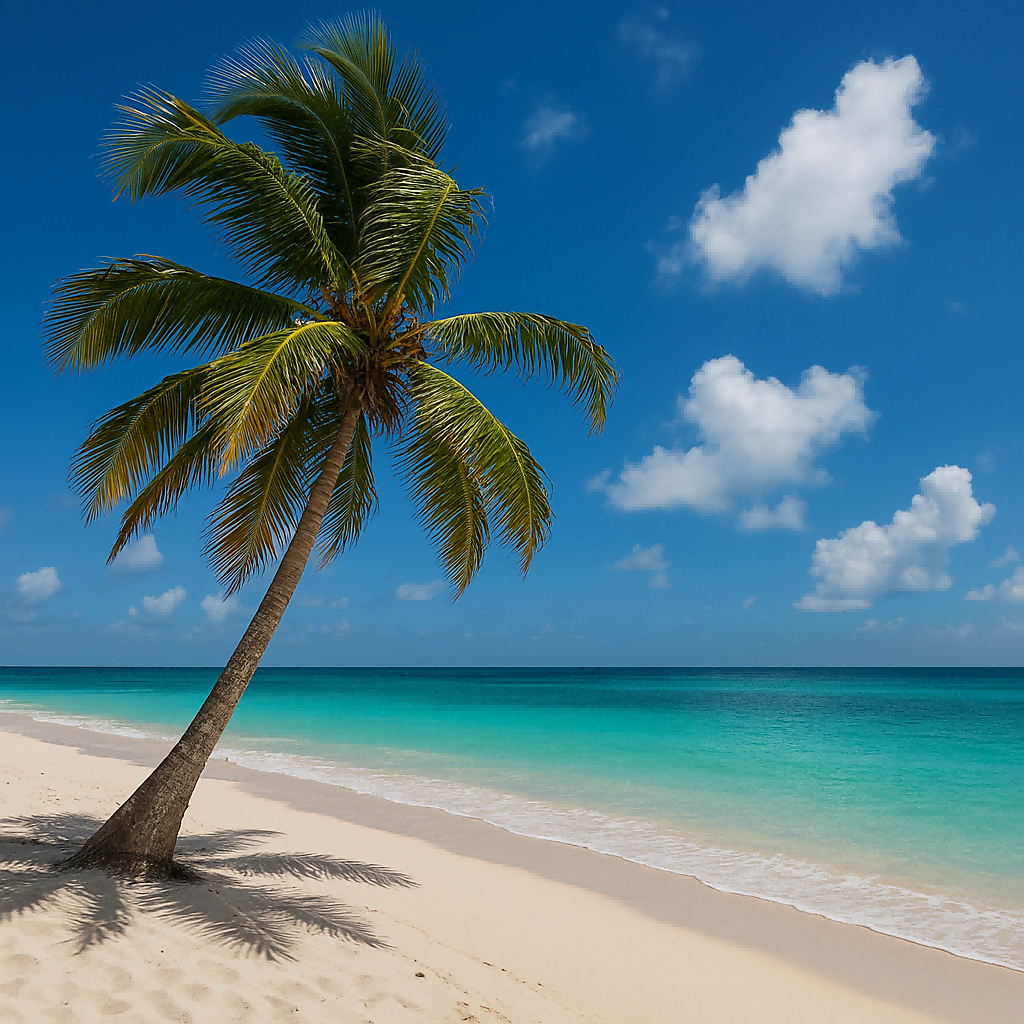}
        {0.00222}
    \end{minipage}\hfill
    \begin{minipage}{0.151\textwidth}
        \centering
        \includegraphics[width=\textwidth]{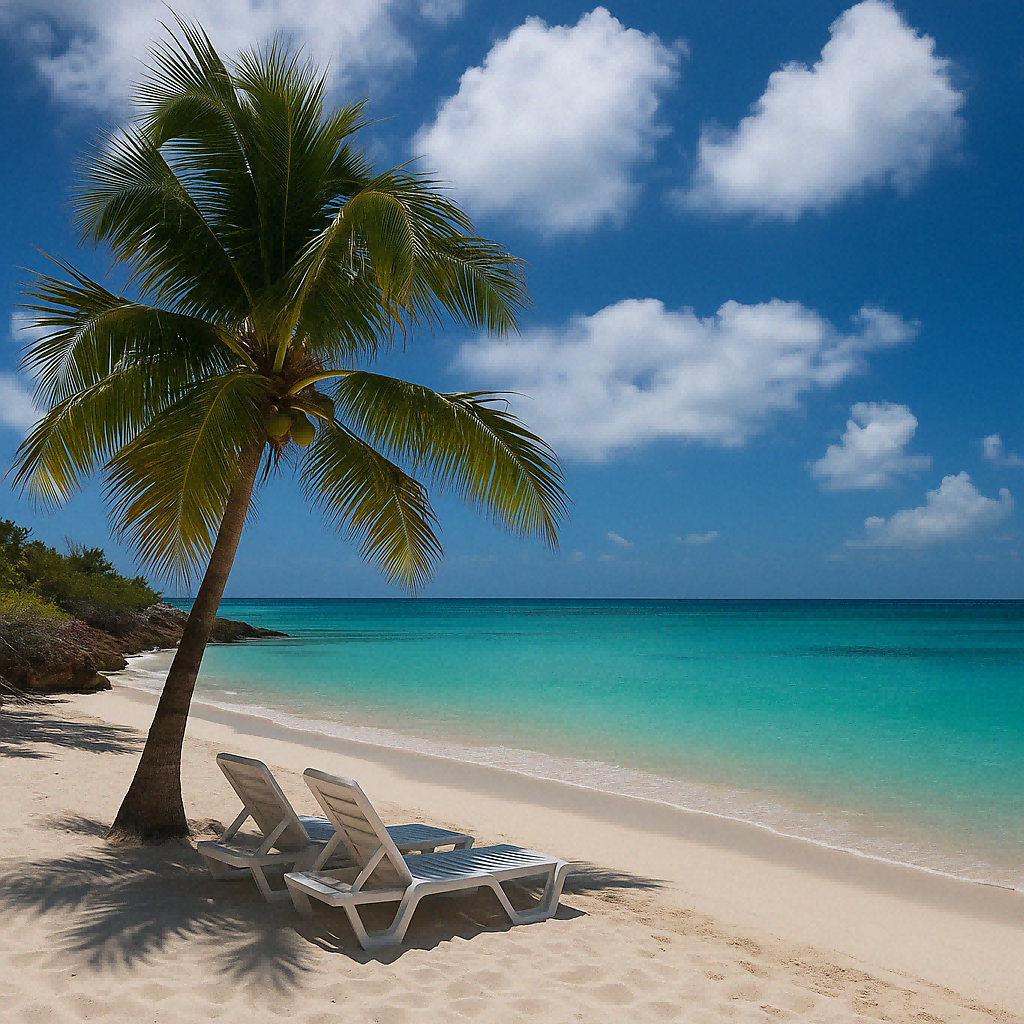}
        {0.00296}
    \end{minipage}\hfill
    \begin{minipage}{0.151\textwidth}
        \centering
        \includegraphics[width=\textwidth]{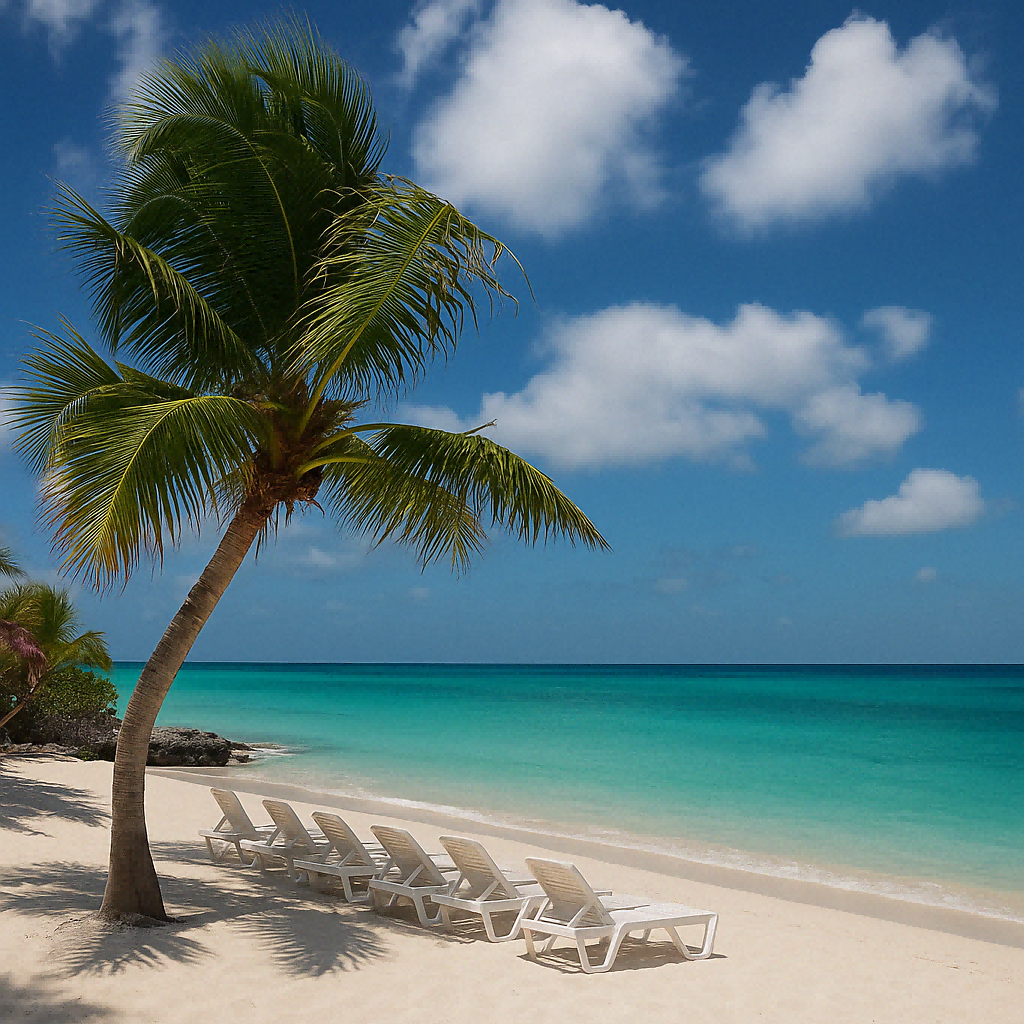}
        {0.00415}
    \end{minipage}

    \begin{minipage}{0.02\textwidth} 
        \vfill
        \centering
        \rotatebox{90}{\textbf{HQ Image}} 
        \vfill
    \end{minipage}%
    \hfill
    \begin{minipage}{0.151\textwidth} 
        \centering
        \includegraphics[width=\textwidth]{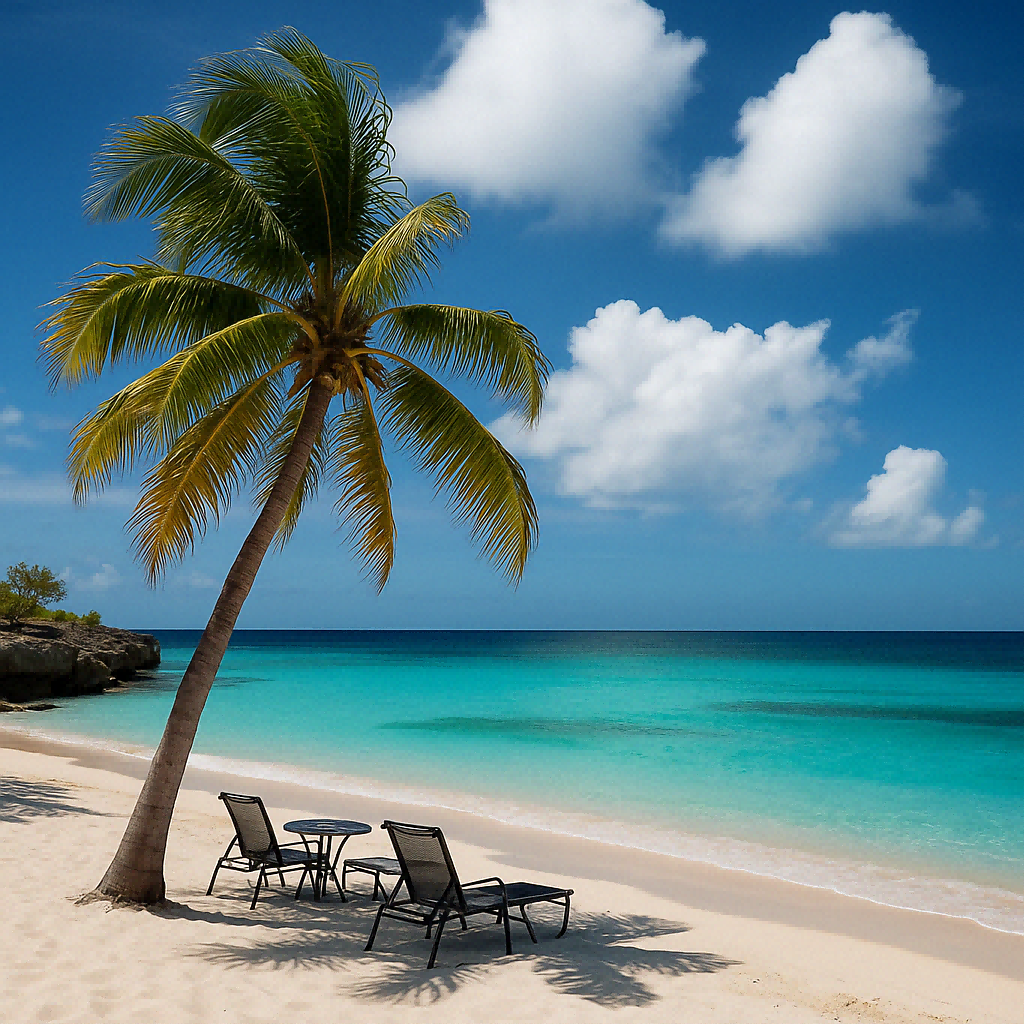}
        {0.00460}
    \end{minipage}\hfill
    \begin{minipage}{0.151\textwidth}
        \centering
        \includegraphics[width=\textwidth]{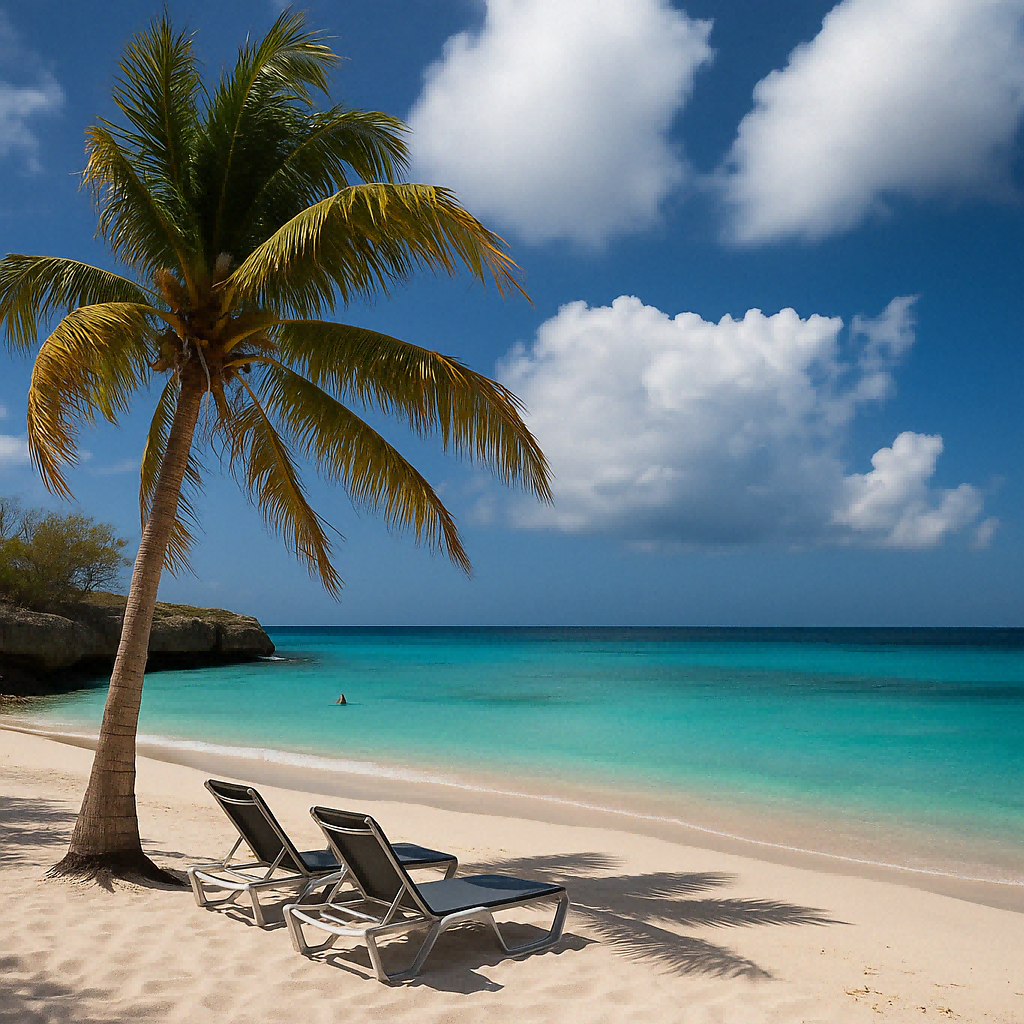}
        {0.00533}
    \end{minipage}\hfill
    \begin{minipage}{0.151\textwidth}
        \centering
        \includegraphics[width=\textwidth]{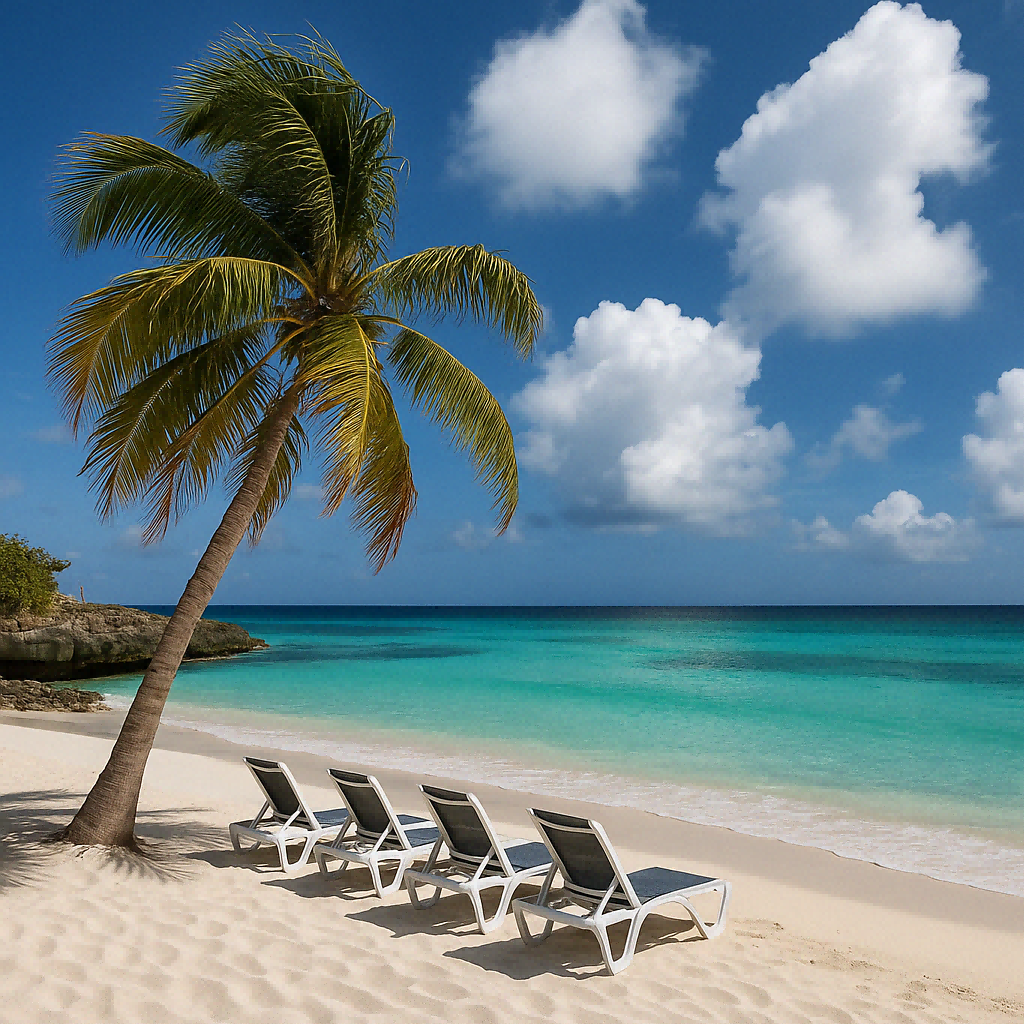}
        {0.00652}
    \end{minipage}
    
    \caption{We present a comparison of compression results based on the same image, using different combinations of text and visual information at varying proportions.“LQ Image” and “HQ Image” refer to the downsampled images encoded at low and high bitrates, respectively. The original image is placed in the top-left corner, and the bitrate required for compression is displayed below each image.}
    \label{fig:qualitative_condition_comparison}
\end{figure}

Based on the same input image, we performed compression using varying proportions of textual and visual information, and present the results in Fig~\ref{fig:qualitative_condition_comparison}. From this series of subjective comparisons, we highlight two generalizable experimental observations: 
First, although it consumes more bitrate, decoding based purely on visual information yields better structural consistency with the original image compared to pure text-based decoding. Due to the limited capacity of text as an information carrier, even allocating more bitrate to text alone often fails to ensure strong consistency. 
Second, when visual information serves as the foundation, a small amount of text can provide complementary cues. However, excessive textual input may negatively impact the quality of the decoded image, as it can introduce unintended modifications. For instance, in the presented case, overly detailed textual descriptions result in the generation of an excessive number of beach chairs.

\subsection{Ablations}
\begin{figure}[tb]
\centerline{\includegraphics[width=0.97\linewidth]{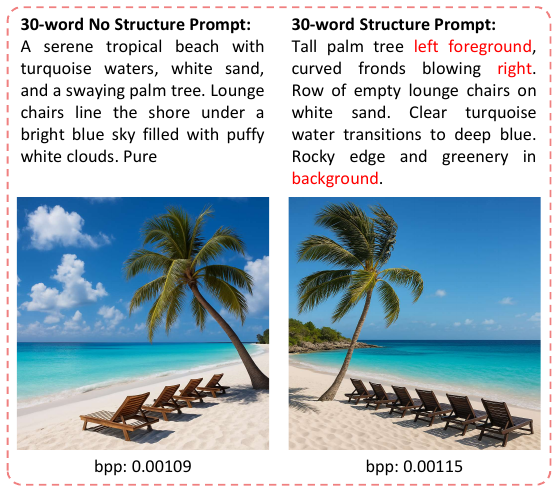}}
    \caption{Image Comparison for Ablation Study of Structure Raster-scan Prompt.}
    \vspace{-5mm}
\label{fig:ablation_results}    
\end{figure}

\myparagraph{Structure Raster-scan Prompt}\quad
\begin{table}[t]
    \centering
    \resizebox{1.0\linewidth}{!}{
    \begin{tabular}{c|l|cc|cc}
        \toprule
        \multirow{2}{*}{\centering Method} & \multirow{2}{*}{\centering Bpp}
        & \multicolumn{2}{c|}{Perception} 
        & \multicolumn{2}{c}{Consistency} \\ \cmidrule(lr){3-4} \cmidrule(lr){5-6}
        & & CLIP-IQA $\uparrow$ & MUSIQ $\uparrow$ & CLIPSIM $\uparrow$ & DISTS $\downarrow$ \\
        \midrule 
        \multirow{2}{*}{Ours} & 0.0008 & \textbf{0.840} & 75.010 & \textbf{0.901} & \textbf{0.289} \\
        & 0.0012 & \textbf{0.856} & \textbf{76.429} & \textbf{0.912} & \textbf{0.274} \\
        \midrule
        \multirow{2}{*}{Ours w/o structure raster-scan prompt} & 0.0007 & 0.818 & \textbf{75.751} & 0.888 & 0.299 \\
        & 0.0011 & 0.847 & 76.115 & 0.902 & 0.281 \\
        \bottomrule
    \end{tabular}}
    \caption{Ablation on our structure raster-scan prompt.}
    \label{tab:ablation_prompt}
\end{table}
As shown in Tab.~\ref{tab:ablation_prompt}, under the text transmission scenario, our structure raster-scan prompt enables better consistency between the reconstructed and original images under comparable bitrates.
The subjective comparison results are shown in Fig.~\ref{fig:ablation_results}.
In this example, we constrain the descriptions to 30 words, generating both unstructured and structured versions.
The structured description includes spatial terms such as ``left foreground," ``right," and ``background," which help the generated images better preserve the structural consistency of the original scene.
In contrast, the unstructured description fails to maintain spatial relationships, resulting in images with left-right positional reversals.

\myparagraph{Prompt Length}\quad
For only textual coding, as shown in Fig.~\ref{fig:main_results}, we test on 15, 30 and 120 words. The word number represents the upper limit of caption length. We observe that from 15 to 30 words, all metrics improve significantly. However, further increasing the number of words does not lead to noticeable gains, suggesting that there is a saturation point beyond which additional textual information brings limited benefit.

For multimodal coding (text + image), we test on 0, 15 and 60 words as in Fig.~\ref{fig:Ablation_prompt_length}. We find that incorporating a short textual prompt (15 words) alongside the image significantly boosts performance across all metrics, especially the two perceptual metrics. Further increasing the prompt length to 60 words improves semantic consistency, but severely degrades structural consistency.

\section{Conclusion}
This work explores a multi-modal image compression framework powered by GPT-4o's image generation capabilities. We investigate two paradigms: text-based and text + image-based compression, aiming to faithfully reconstruct the original image where most or all pixel-level information is regenerated. To better preserve semantic and structural consistency, we introduce a structured raster-scan prompt strategy, asking GPT-4o to list items from top to bottom and left to right, followed by details like style and photometric information. Experiments demonstrate that our method achieves competitive performance compared to recent generative compression approaches at ultra-low bitrates, even without any additional training, highlighting the potential of AIGC in image compression.



{
    \small
    \bibliographystyle{ieeenat_fullname}
    \bibliography{main}

\begin{thebibliography}{58}
\providecommand{\natexlab}[1]{#1}
\providecommand{\url}[1]{\texttt{#1}}
\expandafter\ifx\csname urlstyle\endcsname\relax
  \providecommand{\doi}[1]{doi: #1}\else
  \providecommand{\doi}{doi: \begingroup \urlstyle{rm}\Url}\fi

\bibitem[zli()]{zlib}
zlib — compression compatible with gzip.
\newblock \url{https://docs.python.org/3/library/zlib.html}.

\bibitem[Agustsson and Timofte(2017)]{div2k}
Eirikur Agustsson and Radu Timofte.
\newblock Ntire 2017 challenge on single image super-resolution: Dataset and study.
\newblock In \emph{Proceedings of the IEEE conference on computer vision and pattern recognition workshops}, pages 126--135, 2017.

\bibitem[Ball{\'e} et~al.(2018)Ball{\'e}, Minnen, Singh, Hwang, and Johnston]{balle2018variational}
Johannes Ball{\'e}, David Minnen, Saurabh Singh, Sung~Jin Hwang, and Nick Johnston.
\newblock Variational image compression with a scale hyperprior.
\newblock In \emph{ICLR}, 2018.

\bibitem[Ball{\'e} et~al.(2020)Ball{\'e}, Chou, Minnen, Singh, Johnston, Agustsson, Hwang, and Toderici]{balle2020nonlinear}
Johannes Ball{\'e}, Philip~A Chou, David Minnen, Saurabh Singh, Nick Johnston, Eirikur Agustsson, Sung~Jin Hwang, and George Toderici.
\newblock Nonlinear transform coding.
\newblock \emph{IEEE Journal of Selected Topics in Signal Processing}, 15\penalty0 (2):\penalty0 339--353, 2020.

\bibitem[Bellard()]{bellard_bpg}
Fabrice Bellard.
\newblock Better portable graphics (bpg) image format.
\newblock \url{https://bellard.org/bpg/}.
\newblock Accessed: 2024-05-22.

\bibitem[Betker et~al.(2023{\natexlab{a}})Betker, Goh, Jing, Brooks, Wang, Li, Ouyang, Zhuang, Lee, Guo, et~al.]{betker2023improving}
James Betker, Gabriel Goh, Li Jing, Tim Brooks, Jianfeng Wang, Linjie Li, Long Ouyang, Juntang Zhuang, Joyce Lee, Yufei Guo, et~al.
\newblock Improving image generation with better captions.
\newblock \emph{Computer Science. https://cdn. openai. com/papers/dall-e-3. pdf}, 2\penalty0 (3):\penalty0 8, 2023{\natexlab{a}}.

\bibitem[Betker et~al.(2023{\natexlab{b}})Betker, Goh, Jing, Brooks, Wang, Li, Ouyang, Zhuang, Lee, Guo, et~al.]{dalle3}
James Betker, Gabriel Goh, Li Jing, Tim Brooks, Jianfeng Wang, Linjie Li, Long Ouyang, Juntang Zhuang, Joyce Lee, Yufei Guo, et~al.
\newblock Improving image generation with better captions.
\newblock \emph{Computer Science. https://cdn. openai. com/papers/dall-e-3. pdf}, 2\penalty0 (3):\penalty0 8, 2023{\natexlab{b}}.

\bibitem[Bross et~al.(2021)Bross, Wang, Ye, Liu, Chen, Sullivan, and Ohm]{vvc}
Benjamin Bross, Ye-Kui Wang, Yan Ye, Shan Liu, Jianle Chen, Gary~J Sullivan, and Jens-Rainer Ohm.
\newblock Overview of the versatile video coding (vvc) standard and its applications.
\newblock \emph{TCSVT}, 2021.

\bibitem[Careil et~al.(2023)Careil, Muckley, Verbeek, and Lathuili{\`e}re]{perco}
Marl{\`e}ne Careil, Matthew~J Muckley, Jakob Verbeek, and St{\'e}phane Lathuili{\`e}re.
\newblock Towards image compression with perfect realism at ultra-low bitrates.
\newblock In \emph{The Twelfth International Conference on Learning Representations}, 2023.

\bibitem[Cheng et~al.(2020)Cheng, Sun, Takeuchi, and Katto]{cheng2020learned}
Zhengxue Cheng, Heming Sun, Masaru Takeuchi, and Jiro Katto.
\newblock Learned image compression with discretized gaussian mixture likelihoods and attention modules.
\newblock In \emph{CVPR}, pages 7939--7948, 2020.

\bibitem[Ding et~al.(2020)Ding, Ma, Wang, and Simoncelli]{dists}
Keyan Ding, Kede Ma, Shiqi Wang, and Eero~P Simoncelli.
\newblock Image quality assessment: Unifying structure and texture similarity.
\newblock \emph{IEEE transactions on pattern analysis and machine intelligence}, 44\penalty0 (5):\penalty0 2567--2581, 2020.

\bibitem[Epstein et~al.(2023)Epstein, Jabri, Poole, Efros, and Holynski]{epstein2023diffusion}
Dave Epstein, Allan Jabri, Ben Poole, Alexei Efros, and Aleksander Holynski.
\newblock Diffusion self-guidance for controllable image generation.
\newblock \emph{Advances in Neural Information Processing Systems}, 36:\penalty0 16222--16239, 2023.

\bibitem[Esser et~al.(2021)Esser, Rombach, and Ommer]{vqgan}
Patrick Esser, Robin Rombach, and Bjorn Ommer.
\newblock Taming transformers for high-resolution image synthesis.
\newblock In \emph{Proceedings of the IEEE/CVF conference on computer vision and pattern recognition}, pages 12873--12883, 2021.

\bibitem[Esser et~al.(2024)Esser, Kulal, Blattmann, Entezari, M{\"u}ller, Saini, Levi, Lorenz, Sauer, Boesel, et~al.]{esser2024scaling}
Patrick Esser, Sumith Kulal, Andreas Blattmann, Rahim Entezari, Jonas M{\"u}ller, Harry Saini, Yam Levi, Dominik Lorenz, Axel Sauer, Frederic Boesel, et~al.
\newblock Scaling rectified flow transformers for high-resolution image synthesis.
\newblock In \emph{Forty-first international conference on machine learning}, 2024.

\bibitem[Feng et~al.(2025)Feng, Ma, Wang, Qi, Chen, Chen, and Wang]{feng2025dit4edit}
Kunyu Feng, Yue Ma, Bingyuan Wang, Chenyang Qi, Haozhe Chen, Qifeng Chen, and Zeyu Wang.
\newblock Dit4edit: Diffusion transformer for image editing.
\newblock In \emph{Proceedings of the AAAI Conference on Artificial Intelligence}, pages 2969--2977, 2025.

\bibitem[Feng et~al.(2023)Feng, Guo, Li, and Chen]{feng2023nvtc}
Runsen Feng, Zongyu Guo, Weiping Li, and Zhibo Chen.
\newblock Nvtc: Nonlinear vector transform coding.
\newblock In \emph{Proceedings of the IEEE/CVF Conference on Computer Vision and Pattern Recognition}, pages 6101--6110, 2023.

\bibitem[Gao et~al.(2024)Gao, Li, Pan, Feng, Guo, Lu, Ren, and Chen]{gao2024unimic}
Yixin Gao, Xin Li, Xiaohan Pan, Runsen Feng, Zongyu Guo, Yiting Lu, Yulin Ren, and Zhibo Chen.
\newblock Unimic: Towards universal multi-modality perceptual image compression.
\newblock \emph{arXiv preprint arXiv:2412.04912}, 2024.

\bibitem[Ghouse et~al.(2023)Ghouse, Petersen, Wiggers, Xu, and Sautiere]{ghouse2023residual}
Noor~Fathima Ghouse, Jens Petersen, Auke Wiggers, Tianlin Xu, and Guillaume Sautiere.
\newblock A residual diffusion model for high perceptual quality codec augmentation.
\newblock \emph{arXiv preprint arXiv:2301.05489}, 2023.

\bibitem[Goodfellow et~al.(2014)Goodfellow, Pouget-Abadie, Mirza, Xu, Warde-Farley, Ozair, Courville, and Bengio]{goodfellow2014generative}
Ian Goodfellow, Jean Pouget-Abadie, Mehdi Mirza, Bing Xu, David Warde-Farley, Sherjil Ozair, Aaron Courville, and Yoshua Bengio.
\newblock Generative adversarial nets.
\newblock \emph{NeurIPS}, 27, 2014.

\bibitem[Guo et~al.(2021)Guo, Zhang, Feng, and Chen]{guo2021causal}
Zongyu Guo, Zhizheng Zhang, Runsen Feng, and Zhibo Chen.
\newblock Causal contextual prediction for learned image compression.
\newblock \emph{TCSVT}, 32\penalty0 (4):\penalty0 2329--2341, 2021.

\bibitem[He et~al.(2022)He, Yang, Peng, Ma, Qin, and Wang]{he2022elic}
Dailan He, Ziming Yang, Weikun Peng, Rui Ma, Hongwei Qin, and Yan Wang.
\newblock Elic: Efficient learned image compression with unevenly grouped space-channel contextual adaptive coding.
\newblock In \emph{CVPR}, pages 5718--5727, 2022.

\bibitem[Ho et~al.(2020)Ho, Jain, and Abbeel]{ho2020denoising}
Jonathan Ho, Ajay Jain, and Pieter Abbeel.
\newblock Denoising diffusion probabilistic models.
\newblock \emph{Advances in neural information processing systems}, 33:\penalty0 6840--6851, 2020.

\bibitem[Hoogeboom et~al.(2023)Hoogeboom, Agustsson, Mentzer, Versari, Toderici, and Theis]{hoogeboom2023high}
Emiel Hoogeboom, Eirikur Agustsson, Fabian Mentzer, Luca Versari, George Toderici, and Lucas Theis.
\newblock High-fidelity image compression with score-based generative models.
\newblock \emph{arXiv preprint arXiv:2305.18231}, 2023.

\bibitem[Jia et~al.(2024)Jia, Li, Li, Li, and Lu]{jia2024generative}
Zhaoyang Jia, Jiahao Li, Bin Li, Houqiang Li, and Yan Lu.
\newblock Generative latent coding for ultra-low bitrate image compression.
\newblock In \emph{Proceedings of the IEEE/CVF Conference on Computer Vision and Pattern Recognition}, pages 26088--26098, 2024.

\bibitem[Kawar et~al.(2023)Kawar, Zada, Lang, Tov, Chang, Dekel, Mosseri, and Irani]{kawar2023imagic}
Bahjat Kawar, Shiran Zada, Oran Lang, Omer Tov, Huiwen Chang, Tali Dekel, Inbar Mosseri, and Michal Irani.
\newblock Imagic: Text-based real image editing with diffusion models.
\newblock In \emph{Proceedings of the IEEE/CVF conference on computer vision and pattern recognition}, pages 6007--6017, 2023.

\bibitem[Ke et~al.(2021)Ke, Wang, Wang, Milanfar, and Yang]{ke2021musiq}
Junjie Ke, Qifei Wang, Yilin Wang, Peyman Milanfar, and Feng Yang.
\newblock Musiq: Multi-scale image quality transformer.
\newblock In \emph{Proceedings of the IEEE/CVF international conference on computer vision}, pages 5148--5157, 2021.

\bibitem[K{\"o}rber et~al.(2024)K{\"o}rber, Kromer, Siebert, Hauke, Mueller-Gritschneder, and Schuller]{Nikolaiperco}
Nikolai K{\"o}rber, Eduard Kromer, Andreas Siebert, Sascha Hauke, Daniel Mueller-Gritschneder, and Bj{\"o}rn Schuller.
\newblock Perco ({SD}): Open perceptual compression.
\newblock In \emph{Workshop on Machine Learning and Compression, NeurIPS 2024}, 2024.

\bibitem[Labs(2024)]{flux2024}
Black~Forest Labs.
\newblock Flux.
\newblock \url{https://github.com/black-forest-labs/flux}, 2024.

\bibitem[Lei et~al.(2023)Lei, Uslu, Hassani, and Bidokhti]{lei2023text}
Eric Lei, Yigit~Berkay Uslu, Hamed Hassani, and Shirin~Saeedi Bidokhti.
\newblock Text + sketch: Image compression at ultra low rates.
\newblock In \emph{ICML 2023 Workshop Neural Compression: From Information Theory to Applications}, 2023.

\bibitem[Li et~al.(2024{\natexlab{a}})Li, Lu, Feng, Wu, Zhang, Liu, Zhai, Lin, and Zhang]{li2024misc}
Chunyi Li, Guo Lu, Donghui Feng, Haoning Wu, Zicheng Zhang, Xiaohong Liu, Guangtao Zhai, Weisi Lin, and Wenjun Zhang.
\newblock Misc: Ultra-low bitrate image semantic compression driven by large multimodal model.
\newblock \emph{arXiv preprint arXiv:2402.16749}, 2024{\natexlab{a}}.

\bibitem[Li et~al.(2021)Li, Shi, and Chen]{li2021task}
Xin Li, Jun Shi, and Zhibo Chen.
\newblock Task-driven semantic coding via reinforcement learning.
\newblock \emph{TIP}, 2021.

\bibitem[Li et~al.(2023)Li, Ren, Jin, Lan, Wang, Zeng, Wang, and Chen]{li2023diffusion}
Xin Li, Yulin Ren, Xin Jin, Cuiling Lan, Xingrui Wang, Wenjun Zeng, Xinchao Wang, and Zhibo Chen.
\newblock Diffusion models for image restoration and enhancement--a comprehensive survey.
\newblock \emph{arXiv preprint arXiv:2308.09388}, 2023.

\bibitem[Li et~al.(2024{\natexlab{b}})Li, Zhou, Wei, Ge, and Jiang]{diffeic}
Zhiyuan Li, Yanhui Zhou, Hao Wei, Chenyang Ge, and Jingwen Jiang.
\newblock Towards extreme image compression with latent feature guidance and diffusion prior.
\newblock \emph{IEEE Transactions on Circuits and Systems for Video Technology}, 2024{\natexlab{b}}.

\bibitem[Li et~al.(2024{\natexlab{c}})Li, Zhou, Wei, Ge, and Mian]{rdeic}
Zhiyuan Li, Yanhui Zhou, Hao Wei, Chenyang Ge, and Ajmal Mian.
\newblock Diffusion-based extreme image compression with compressed feature initialization.
\newblock \emph{arXiv preprint arXiv:2410.02640}, 2024{\natexlab{c}}.

\bibitem[Liu et~al.(2023)Liu, Sun, and Katto]{liu2023learned}
Jinming Liu, Heming Sun, and Jiro Katto.
\newblock Learned image compression with mixed transformer-cnn architectures.
\newblock In \emph{Proceedings of the IEEE/CVF Conference on Computer Vision and Pattern Recognition}, pages 14388--14397, 2023.

\bibitem[Mao et~al.(2024)Mao, Yang, Zhang, Wang, Wang, Wang, Jin, and Ma]{mao2024extreme}
Qi Mao, Tinghan Yang, Yinuo Zhang, Zijian Wang, Meng Wang, Shiqi Wang, Libiao Jin, and Siwei Ma.
\newblock Extreme image compression using fine-tuned vqgans.
\newblock In \emph{2024 Data Compression Conference (DCC)}, pages 203--212. IEEE, 2024.

\bibitem[Minnen and Singh(2020)]{minnen2020channel}
David Minnen and Saurabh Singh.
\newblock Channel-wise autoregressive entropy models for learned image compression.
\newblock In \emph{2020 IEEE International Conference on Image Processing (ICIP)}, pages 3339--3343. IEEE, 2020.

\bibitem[Minnen et~al.(2018)Minnen, Ball{\'e}, and Toderici]{minnen2018joint}
David Minnen, Johannes Ball{\'e}, and George Toderici.
\newblock Joint autoregressive and hierarchical priors for learned image compression.
\newblock In \emph{NeurIPS}, 2018.

\bibitem[Mokady et~al.(2023)Mokady, Hertz, Aberman, Pritch, and Cohen-Or]{Mokady2023NullText}
Ron Mokady, Amir Hertz, Kfir Aberman, Yael Pritch, and Daniel Cohen-Or.
\newblock Null-text inversion for editing real images using guided diffusion models.
\newblock In \emph{Proceedings of the IEEE/CVF Conference on Computer Vision and Pattern Recognition (CVPR)}, pages 6038--6047, 2023.

\bibitem[Mou et~al.(2024)Mou, Wang, Xie, Wu, Zhang, Qi, and Shan]{mou2024t2i}
Chong Mou, Xintao Wang, Liangbin Xie, Yanze Wu, Jian Zhang, Zhongang Qi, and Ying Shan.
\newblock T2i-adapter: Learning adapters to dig out more controllable ability for text-to-image diffusion models.
\newblock In \emph{Proceedings of the AAAI conference on artificial intelligence}, pages 4296--4304, 2024.

\bibitem[Muckley et~al.(2021)Muckley, Juravsky, Severo, Singh, Duval, and Ullrich]{muckley2021neuralcompression}
Matthew Muckley, Jordan Juravsky, Daniel Severo, Mannat Singh, Quentin Duval, and Karen Ullrich.
\newblock Neuralcompression.
\newblock \url{https://github.com/facebookresearch/NeuralCompression}, 2021.

\bibitem[Muckley et~al.(2023)Muckley, El-Nouby, Ullrich, J{\'e}gou, and Verbeek]{msillm}
Matthew~J Muckley, Alaaeldin El-Nouby, Karen Ullrich, Herv{\'e} J{\'e}gou, and Jakob Verbeek.
\newblock Improving statistical fidelity for neural image compression with implicit local likelihood models.
\newblock In \emph{International Conference on Machine Learning}, pages 25426--25443. PMLR, 2023.

\bibitem[OpenAI(2025)]{openai2025gpt4o}
OpenAI.
\newblock Addendum to gpt-4o system card: 4o image generation, 2025.
\newblock Accessed: 2025-04-22.

\bibitem[Radford et~al.(2021)Radford, Kim, Hallacy, Ramesh, Goh, Agarwal, Sastry, Askell, Mishkin, Clark, et~al.]{clip}
Alec Radford, Jong~Wook Kim, Chris Hallacy, Aditya Ramesh, Gabriel Goh, Sandhini Agarwal, Girish Sastry, Amanda Askell, Pamela Mishkin, Jack Clark, et~al.
\newblock Learning transferable visual models from natural language supervision.
\newblock In \emph{International conference on machine learning}, pages 8748--8763. PMLR, 2021.

\bibitem[Rombach et~al.(2022)Rombach, Blattmann, Lorenz, Esser, and Ommer]{latentdiffusion}
Robin Rombach, Andreas Blattmann, Dominik Lorenz, Patrick Esser, and Bj{\"o}rn Ommer.
\newblock High-resolution image synthesis with latent diffusion models.
\newblock In \emph{Proceedings of the IEEE/CVF conference on computer vision and pattern recognition}, pages 10684--10695, 2022.

\bibitem[Shi et~al.(2024)Shi, Xue, Liew, Pan, Yan, Zhang, Tan, and Bai]{shi2024dragdiffusion}
Yujun Shi, Chuhui Xue, Jun~Hao Liew, Jiachun Pan, Hanshu Yan, Wenqing Zhang, Vincent~YF Tan, and Song Bai.
\newblock Dragdiffusion: Harnessing diffusion models for interactive point-based image editing.
\newblock In \emph{Proceedings of the IEEE/CVF Conference on Computer Vision and Pattern Recognition}, pages 8839--8849, 2024.

\bibitem[Song et~al.(2020)Song, Sohl-Dickstein, Kingma, Kumar, Ermon, and Poole]{song2020score}
Yang Song, Jascha Sohl-Dickstein, Diederik~P Kingma, Abhishek Kumar, Stefano Ermon, and Ben Poole.
\newblock Score-based generative modeling through stochastic differential equations.
\newblock \emph{arXiv preprint arXiv:2011.13456}, 2020.

\bibitem[Wang et~al.(2023{\natexlab{a}})Wang, Chan, and Loy]{clipiqa}
Jianyi Wang, Kelvin~CK Chan, and Chen~Change Loy.
\newblock Exploring clip for assessing the look and feel of images.
\newblock In \emph{Proceedings of the AAAI conference on artificial intelligence}, pages 2555--2563, 2023{\natexlab{a}}.

\bibitem[Wang et~al.(2023{\natexlab{b}})Wang, Zhao, and Xing]{wang2023stylediffusion}
Zhizhong Wang, Lei Zhao, and Wei Xing.
\newblock Stylediffusion: Controllable disentangled style transfer via diffusion models.
\newblock In \emph{Proceedings of the IEEE/CVF International Conference on Computer Vision}, pages 7677--7689, 2023{\natexlab{b}}.

\bibitem[Wu et~al.(2021)Wu, Li, Zhang, Jin, and Chen]{wu2021learned}
Yaojun Wu, Xin Li, Zhizheng Zhang, Xin Jin, and Zhibo Chen.
\newblock Learned block-based hybrid image compression.
\newblock \emph{IEEE Transactions on Circuits and Systems for Video Technology}, 32\penalty0 (6):\penalty0 3978--3990, 2021.

\bibitem[Xie et~al.(2025)Xie, Chen, Zhao, Yu, Zhu, Lin, Zhang, Li, Chen, Cai, et~al.]{xie2025sana}
Enze Xie, Junsong Chen, Yuyang Zhao, Jincheng Yu, Ligeng Zhu, Yujun Lin, Zhekai Zhang, Muyang Li, Junyu Chen, Han Cai, et~al.
\newblock Sana 1.5: Efficient scaling of training-time and inference-time compute in linear diffusion transformer.
\newblock \emph{arXiv preprint arXiv:2501.18427}, 2025.

\bibitem[Xue et~al.(2024)Xue, Mao, Wang, Zhang, and Ma]{xue2024unifying}
Naifu Xue, Qi Mao, Zijian Wang, Yuan Zhang, and Siwei Ma.
\newblock Unifying generation and compression: Ultra-low bitrate image coding via multi-stage transformer.
\newblock In \emph{2024 IEEE International Conference on Multimedia and Expo (ICME)}, pages 1--6. IEEE, 2024.

\bibitem[Xue et~al.(2025)Xue, Jia, Li, Li, Zhang, and Lu]{xue2025dlf}
Naifu Xue, Zhaoyang Jia, Jiahao Li, Bin Li, Yuan Zhang, and Yan Lu.
\newblock Dlf: Extreme image compression with dual-generative latent fusion.
\newblock \emph{arXiv preprint arXiv:2503.01428}, 2025.

\bibitem[Yan et~al.(2025)Yan, Ye, Li, Huang, Yuan, He, Lin, He, He, and Yuan]{gptimgeval}
Zhiyuan Yan, Junyan Ye, Weijia Li, Zilong Huang, Shenghai Yuan, Xiangyang He, Kaiqing Lin, Jun He, Conghui He, and Li Yuan.
\newblock Gpt-imgeval: A comprehensive benchmark for diagnosing gpt4o in image generation.
\newblock \emph{arXiv preprint arXiv:2504.02782}, 2025.

\bibitem[Yang and Mandt(2024)]{yang2024lossy}
Ruihan Yang and Stephan Mandt.
\newblock Lossy image compression with conditional diffusion models.
\newblock \emph{Advances in Neural Information Processing Systems}, 36, 2024.

\bibitem[Zhang et~al.(2023{\natexlab{a}})Zhang, Rao, and Agrawala]{controlnet}
Lvmin Zhang, Anyi Rao, and Maneesh Agrawala.
\newblock Adding conditional control to text-to-image diffusion models.
\newblock In \emph{Proceedings of the IEEE/CVF International Conference on Computer Vision}, pages 3836--3847, 2023{\natexlab{a}}.

\bibitem[Zhang et~al.(2023{\natexlab{b}})Zhang, Huang, Tang, Huang, Ma, Dong, and Xu]{zhang2023inversion}
Yuxin Zhang, Nisha Huang, Fan Tang, Haibin Huang, Chongyang Ma, Weiming Dong, and Changsheng Xu.
\newblock Inversion-based style transfer with diffusion models.
\newblock In \emph{Proceedings of the IEEE/CVF conference on computer vision and pattern recognition}, pages 10146--10156, 2023{\natexlab{b}}.

\bibitem[Zhang et~al.(2024)Zhang, Zhang, Xing, Li, Zhao, Sun, Lan, Luan, Huang, and Lin]{zhang2024artbank}
Zhanjie Zhang, Quanwei Zhang, Wei Xing, Guangyuan Li, Lei Zhao, Jiakai Sun, Zehua Lan, Junsheng Luan, Yiling Huang, and Huaizhong Lin.
\newblock Artbank: Artistic style transfer with pre-trained diffusion model and implicit style prompt bank.
\newblock In \emph{Proceedings of the AAAI Conference on Artificial Intelligence}, pages 7396--7404, 2024.

\end{thebibliography}
}

\clearpage
\appendix
\section{More Experimental Details}
\label{sec:app_expermential_details}
\myparagraph{Implementation Details.} 

For MS-ILLM~\cite{msillm}, we use the publicly released checkpoints in the official repository~\cite{muckley2021neuralcompression}. 

For Test + Sketch~\cite{lei2023text}, we use the code and pretrained checkpoints from \url{https://github.com/leieric/Text-Sketch}.

For Perco~\cite{perco}, we use the implementation based on Stable Diffusion v2.1~\cite{Nikolaiperco} and pretrained checkpoints from \url{https://github.com/Nikolai10/PerCo}.

\myparagraph{Metric Computation}
We use the \texttt{pyiqa} package (available from \url{https://github.com/chaofengc/IQA-PyTorch}) to calculate CLIP-IQA and MUSIQ. We use the \texttt{TorchMetrics} package (available from \url{https://github.com/Lightning-AI/torchmetrics}) to calculate DISTS.

For the calculation of CLIPSIM, we used the \texttt{'ViT-B-32'} model from OpenCLIP package (available from \url{https://github.com/mlfoundations/open_clip}) to extract the clip features of the original and decoded images, and then compute the cosine similarity between these features.


\section{More Qualitative Results}

\begin{figure}[htbp]
    \centering

    \begin{minipage}{0.02\textwidth} 
        \vfill
        \centering
        \rotatebox{90}{\textbf{Text}} 
        \vfill
    \end{minipage}%
    \hfill
    \begin{minipage}{0.15\textwidth} 
        \centering
        \textbf{Sample 1}
        \includegraphics[width=\textwidth]{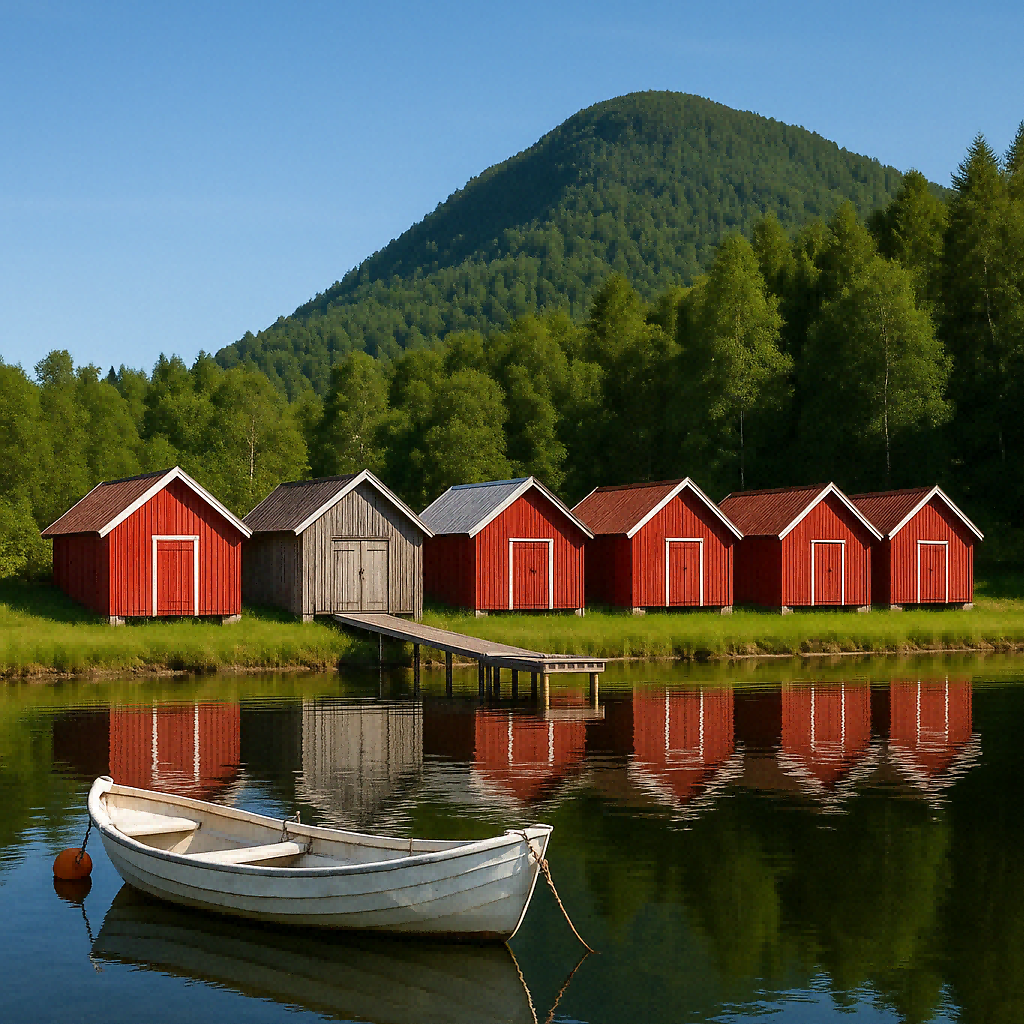} 
    \end{minipage}\hfill
    \begin{minipage}{0.15\textwidth}
        \centering
        \textbf{Sample 2}
        \includegraphics[width=\textwidth]{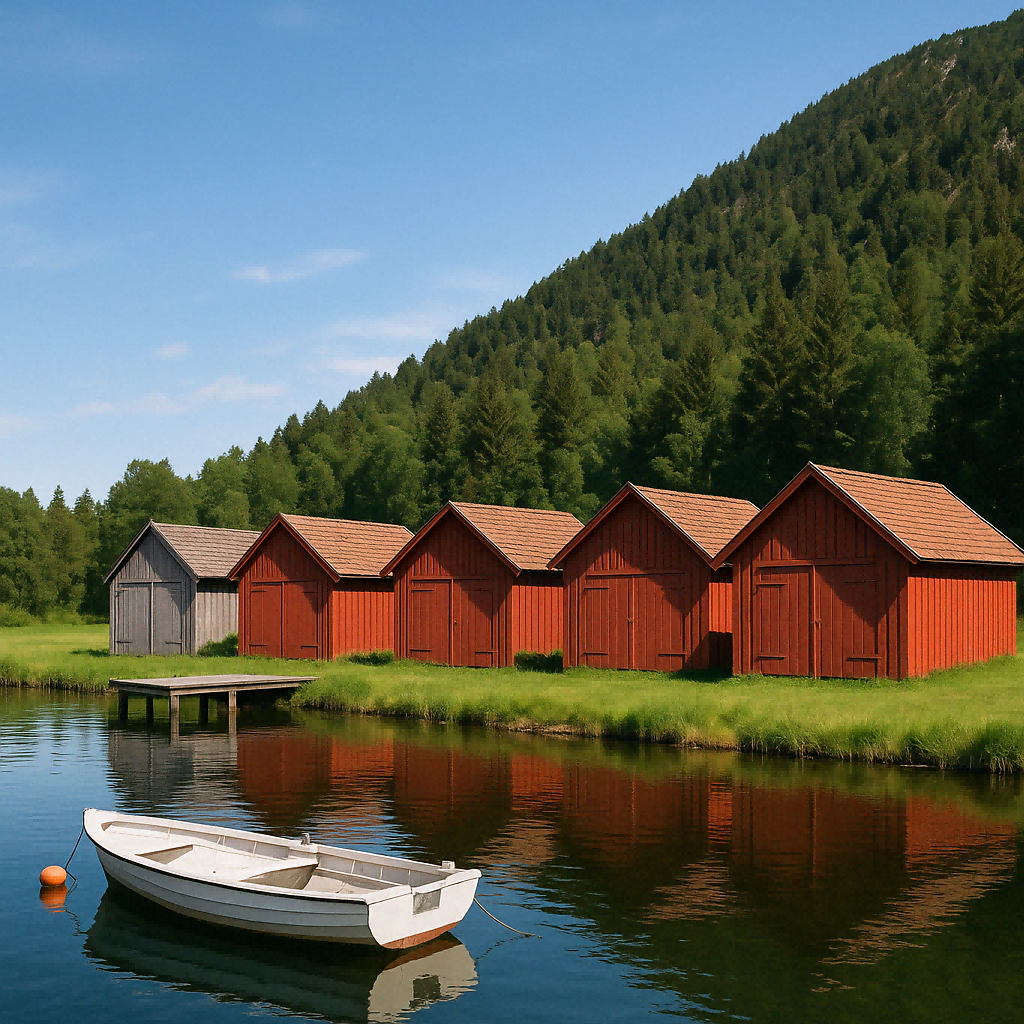}
    \end{minipage}\hfill
    \begin{minipage}{0.15\textwidth}
        \centering
        \textbf{Sample 3}
        \includegraphics[width=\textwidth]{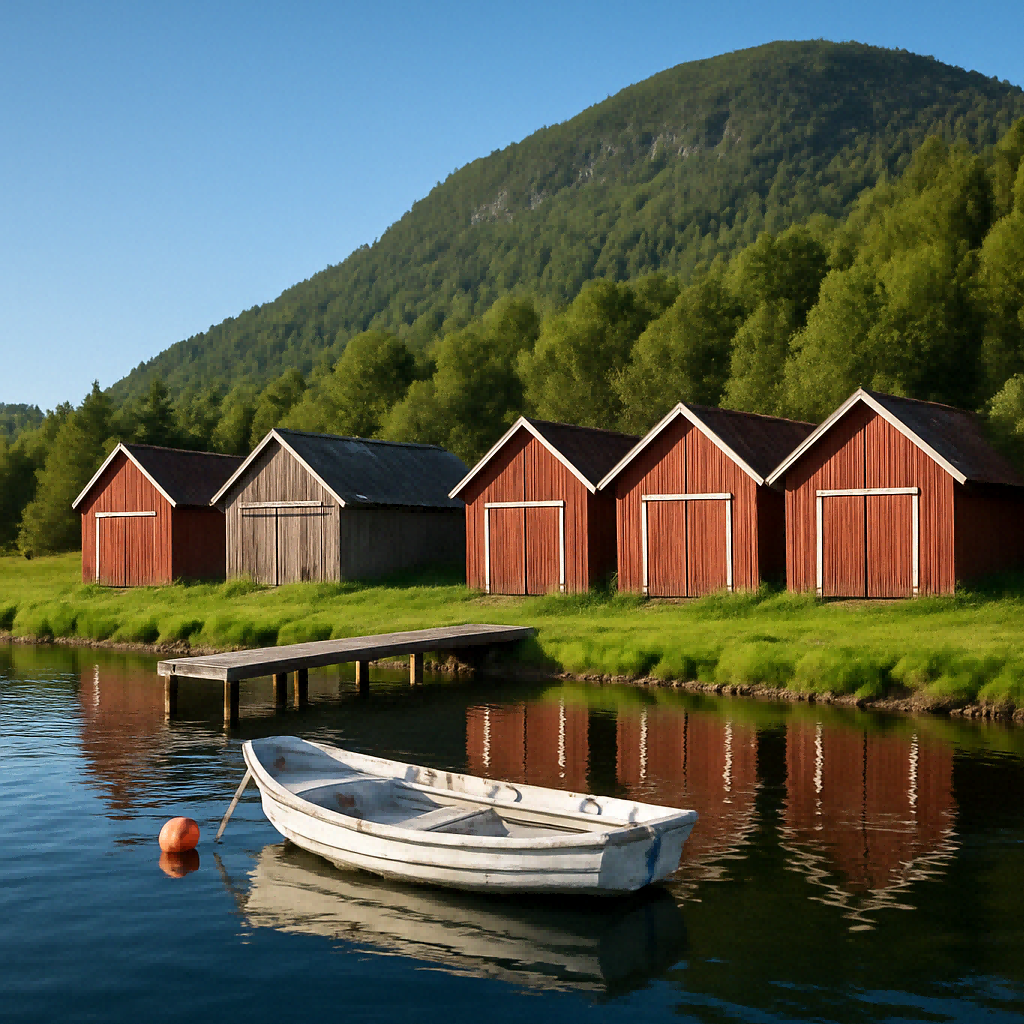}
    \end{minipage}

    \begin{minipage}{0.02\textwidth} 
        \vfill
        \centering
        \rotatebox{90}{\textbf{Text+Image}} 
        \vfill
    \end{minipage}%
    \hfill
    \begin{minipage}{0.15\textwidth} 
        \centering
        \includegraphics[width=\textwidth]{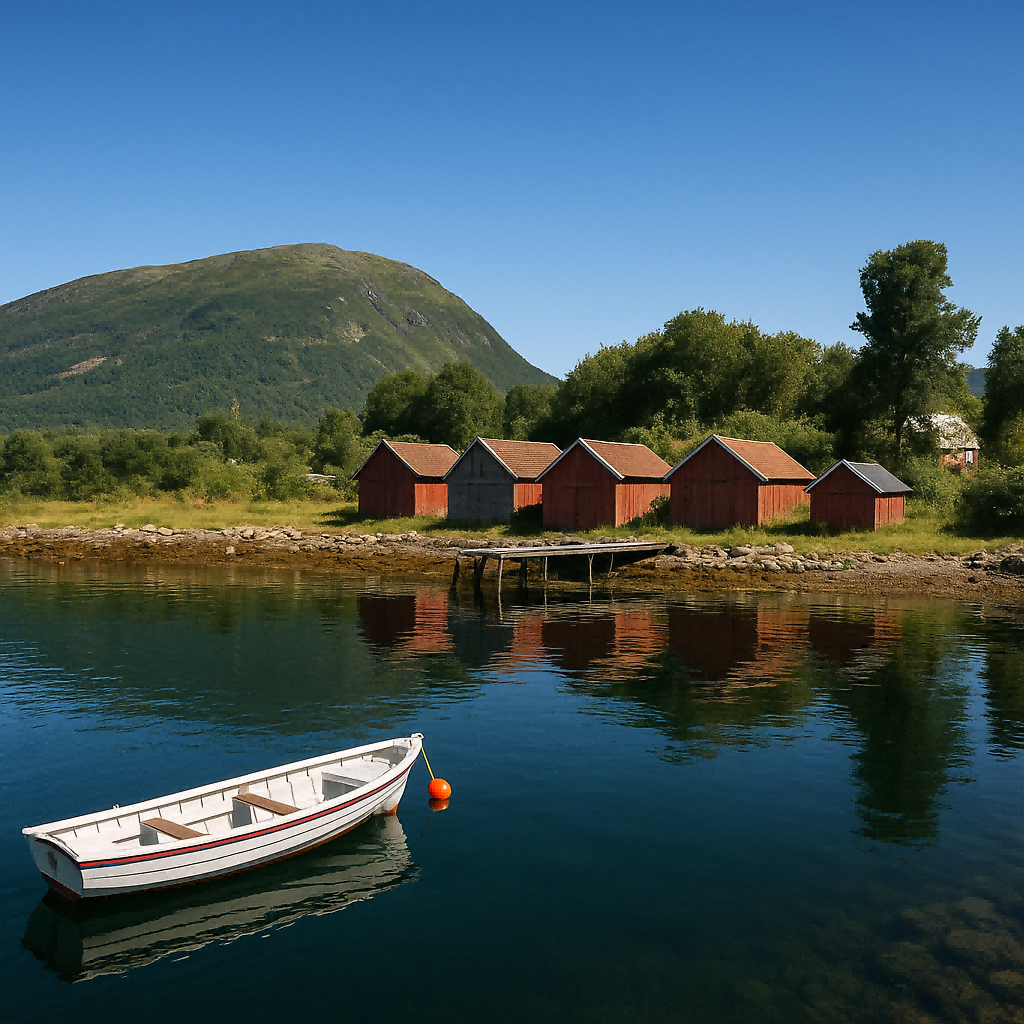} 
    \end{minipage}\hfill
    \begin{minipage}{0.15\textwidth}
        \centering
        \includegraphics[width=\textwidth]{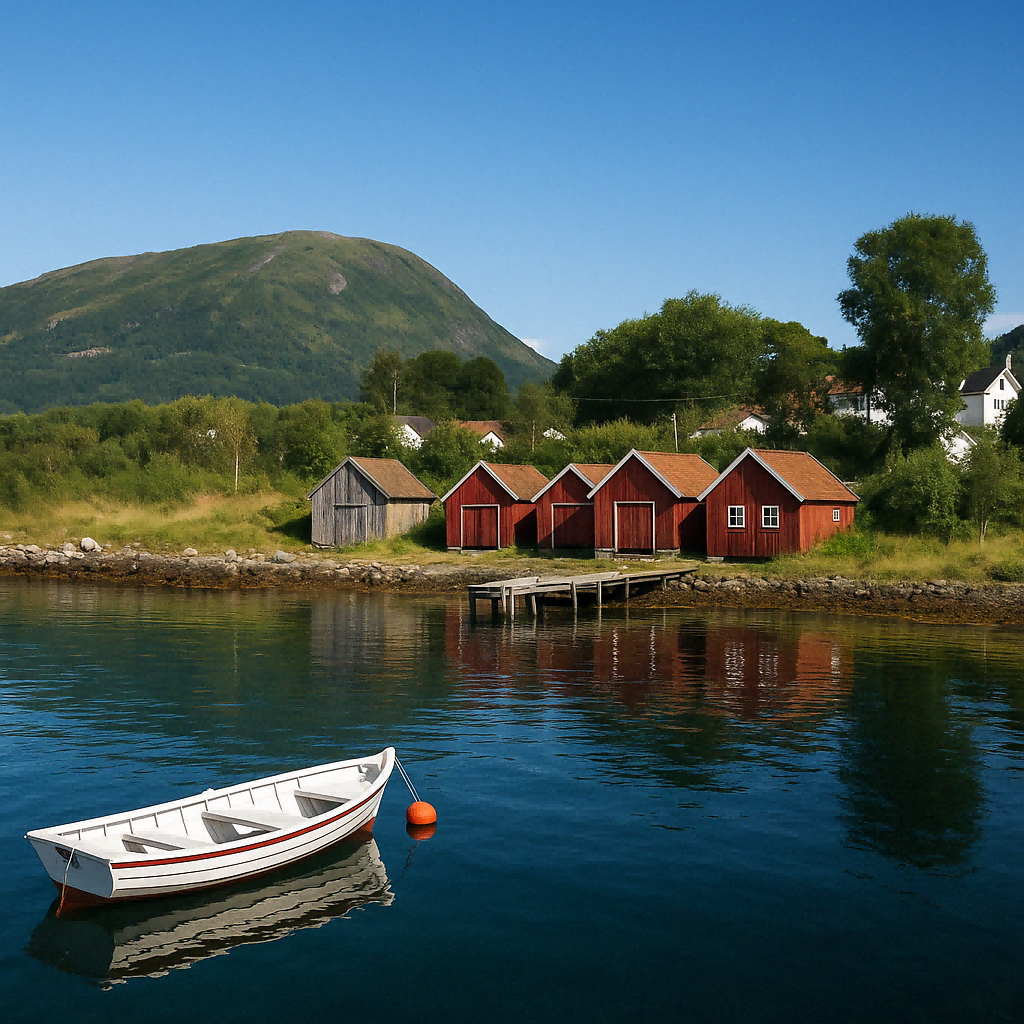}
    \end{minipage}\hfill
    \begin{minipage}{0.15\textwidth}
        \centering
        \includegraphics[width=\textwidth]{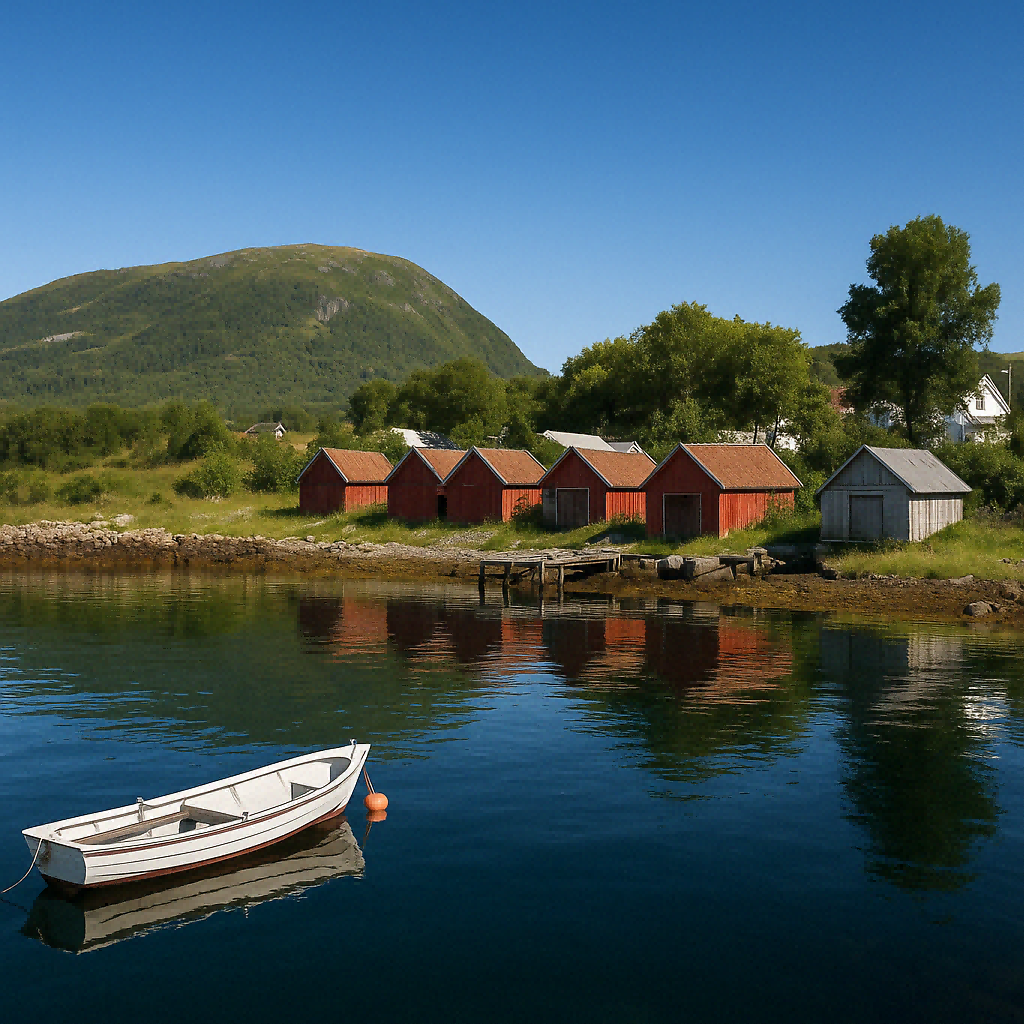}
    \end{minipage}
    
    \caption{We present results generated multiple times at the decoding side using the same transmitted information. Compared to pure text, the text+image setting produces more stable outputs and is less susceptible to randomness.}
    \label{fig:multi_try}
\end{figure}
\end{document}